\documentclass[sigconf]{acmart}

\usepackage{subfigure}
\usepackage{booktabs}
\usepackage{algorithm}
\usepackage{algorithmic}
\usepackage{multirow}
\usepackage{color} 
\def\BibTeX{{\rm B\kern-.05em{\sc i\kern-.025em b}\kern-.08emT\kern-.1667em\lower.7ex\hbox{E}\kern-.125emX}}
    
\begin{document}

\title{Domain-adversarial Network Alignment}
\author{Huiting Hong}
\affiliation{\institution{School of Computer Science}Beijing Institute of Technology, China}
\email{whitneyhung1993@gmail.com}
\author{Xin Li$^*$}
\affiliation{\institution{School of Computer Science}Beijing Institute of Technology, China}
\email{xinli@bit.edu.cn}
\author{Yuangang Pan}
\affiliation{University of Technology Sydney, Australia}
\email{Yuangang.Pan@student.uts.edu.au}
\author{Ivor Tsang}
\affiliation{University of Technology Sydney, Australia}
\email{Ivor.Tsang@uts.edu.au}


\begin{abstract}
Network alignment is a critical task to a wide variety of fields. Many existing works leverage on representation learning to accomplish this task without eliminating domain representation bias induced by domain-dependent features, which yield inferior alignment performance. This paper proposes a unified deep architecture (\textit{DANA}) to obtain a domain-invariant representation for network alignment via an adversarial domain classifier. Specifically, we employ the graph convolutional networks to perform network embedding under the domain adversarial principle, given a small set of observed anchors. Then, the semi-supervised learning framework is optimized by maximizing a posterior probability distribution of observed anchors and the loss of a domain classifier simultaneously. We also develop a few variants of our model, such as, direction-aware network alignment, weight-sharing for directed networks and simplification of parameter space. Experiments on three real-world social network datasets demonstrate that our proposed approaches achieve state-of-the-art alignment results.


%






\end{abstract}



%
\maketitle
\section{Introduction}

Network alignment seeks to find the correspondence of nodes (a.k.a. anchor links) across two or more networks. It is of importance in a wide variety of fields. For instance, network alignment can be applied to connecting identical users across different social network medias (refer to as different domains in the sequel). The established user correspondence could alleviate the sparsity issue of analyzing individual social networks with information fusion, benefiting applications such as preferred link prediction and cross-domain recommendation. Similarly, network alignment can help construct a more compact knowledge graph based on the existing vertical or cross-lingual knowledge bases, thus to obtain better knowledge inference. In Bioinformatics, aligning protein-protein interaction networks from different species has been widely studied in order to determine the common functional structures.

Regarding the network alignment task, there exists a basic assumption that affiliated nodes should have a consistent connectivity structure across the different networks. The approaches exploring the topological consistency offer a universal solution to the alignment task, since the informative node attributes are usually unavailable in reality. Recently, representation learning of networks a.k.a. network embedding has provided a means to obtain low-dimensional representations of nodes by exploiting the structural information of the network. Then, the network alignment could be performed by exploring a common low-dimensional subspace of networks or a subspace transformation between networks.

\begin{figure}[t]
\centering
    \subfigure[SNNA]{
        \includegraphics[width=0.215\textwidth]{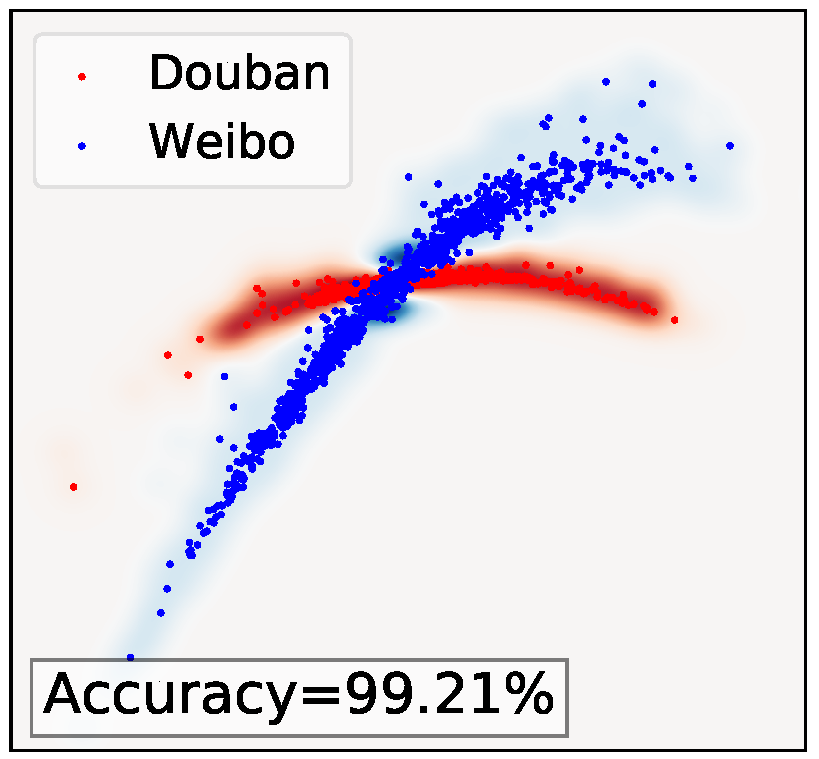}
        \label{fig:v_SNNAu}
    }
    \subfigure[IONE]{
        \includegraphics[width=0.245\textwidth]{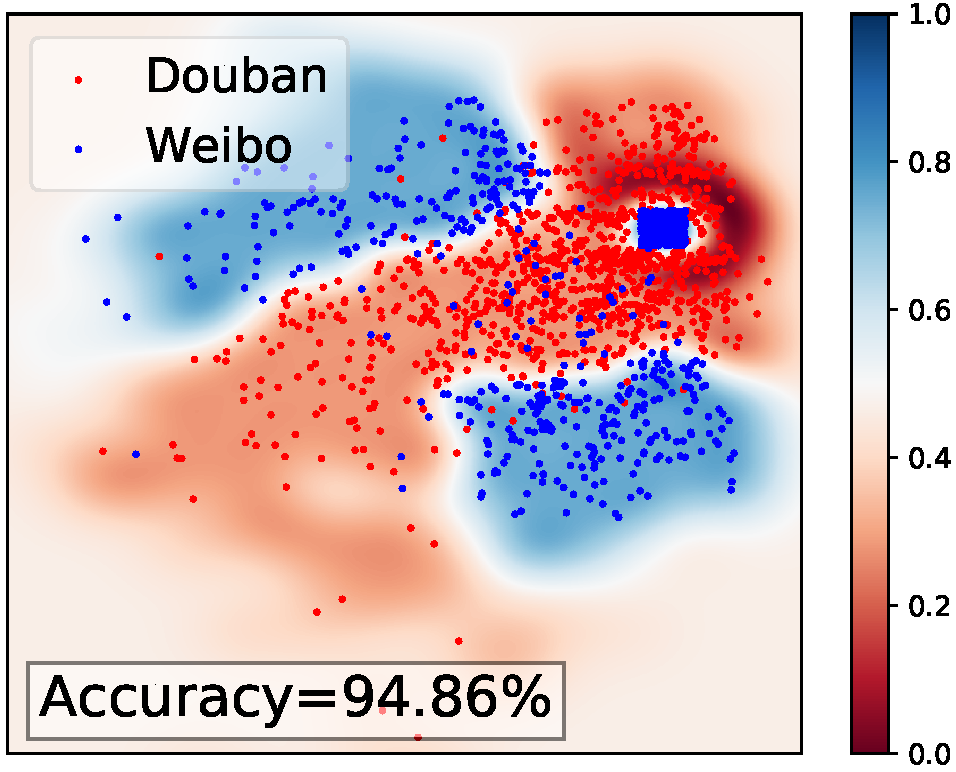}
        \label{fig:v_IONE}
    }
\caption{\label{fig:visualization-introduction}An SVM trained domain classification on 2D representations of vertices obtained by existing alignment approaches in Douban-weibo dataset.}
\end{figure}

However, in the literature, existing embedding-based alignment methods, e.g. \textit{SNNA} \cite{SNNA} and \textit{IONE} \cite{liu2016aligning}, fail to explicitly capture domain-invariant features, which therefore suffer from domain representation bias w.r.t. the network alignment task\footnote{In this paper, the domain representation bias refers to the domain-dependent features which are irrelevant to the specific task but is able to represent domains. For example, RGB value could be thte key feature to distingish from colorful digits and grayscale digits, but shouldn't be the key feature to disignuish from each digit.}. Most network-embedding approaches tend to obtain the local structures and high-order structures simultaneously in the embedded space. For example, \textit{IONE} leveraged \textit{LINE} \cite{tang2015line} to preserve the second-order proximity explicitly and retain high-order structures implicitly via linkage propagation. The learned embedding therefore includes domain-dependent signals, which may be suitable for distinguishing between the domains/networks, but is inborn defective for the alignment task due to inadequate learning of domain-invariant features.



Fig.\ref{fig:v_SNNAu} and \ref{fig:v_IONE} show the 2D representations of nodes of two networks (Douban and Weibo), which are obtained from two state-of-the-art network alignment approaches \textit{SNNA}\cite{SNNA} and \textit{IONE}\cite{liu2016aligning} respectively. For clarity, we only plot 2000 vertices randomly sampled from the test set. The experimental setup is consistent with that described in Sec.4. The decision boundaries of SVM is shown in the background color. The \textit{SVM} {domain} classifiers are trained on the learned representations and the testing accuracies are 0.99 and 0.95 respectively. We believe that the representations somehow encoded the domain-dependent feature, for example, the signal of the average node degree (the average node degree of Douban is twice that of Weibo, see Table \ref{tab:dataset}). And we argue that such domain-dependent features learned by existing network alignment approaches are not informative to align the networks, as the domain of each network is previously known to the alignment task. And sometimes the domain-dependent features may even lead to an inferior alignment performance. Thus, suppressing the learning of domain-dependent features/domain representation bias to lead the representations of nodes more task-specific to boost the alignment performance is the basic motivation in this paper.

In the literature, there {are some} existing works which introduce domain-dependent features and domain-independent features in pursuit of better performance for cross-domain tasks, e.g., cross-domain sentiment analysis and image segmentation \cite{weiss2016survey}. These features are usually learned through manual selection or (and) feature augmentation, which is applicable in the field of natural language processing and image processing, where explicit semantics and rich attributes are accessible \cite{panjialinWWW10}. However, it cannot be applied to network embedding, where only structural information is available.




Inspired by the recent advancement of domain adaptation learning \cite{ganin2016domain,xie2017controllable}, which is trying to obtain features that are invariant to the change of domains, we propose to incorporate an adversarial learning of domain classifier into the process of network embedding within an alignment framework to suppress the generation of the domain-dependent features for better alignment performance. The framework - \textbf{D}omain-\textbf{A}dversarial \textbf{N}etwork \textbf{A}lignment (DANA) mainly consists of two components, namely, task-driven network embedding module and adversarial domain classifier. 

In this paper, the task-driven embedding of networks is accomplished via graph convolutional networks (GCNs) \cite{kipf2016semi,defferrard2016convolutional}, known as being powerful on graph-structured data. Instead of enforcing the anchors' representations to be same as in most existing works, e.g., \textit{IONE}, we maximize a posterior probability distribution of anchors over the parameter space to supervise \textit{GCNs} in pursuit of a more flexible network representation. On the other hand, the embedding process is also supervised by the adversarial  domain classifier, which is meant to perform an adversarial learning of the domain classifier to obtain the domain-invariant features w.r.t. the alignment task. That is to say, the framework is optimized in order to minimize the loss of the alignment and maximize the loss of the domain classifier simultaneously. 

To better deal with the alignment task involved with directed networks, e.g., Twitter where follower-followee relations\footnote{In twitter, someone is following you does not mean that you are necessarily following them back. In contrast, the friendship on Facebook is always bidirectional, meaning  that the contact graph is undirected.} are maintained on purpose in Twitter to constitute a directed network/graph, we further adapt the framework by developing a direction-aware structure to characterize the directed edges in networks. Moreover, weight-sharing within the network embedding module is facilitated to obtain similar subspaces for each domain/network, which generally benefits the alignment determination, while reducing the number of parameters to speed up the training process. a t

The main contributions of this paper can be summarized as follows:
\begin{itemize}
    \item We propose a representation learning-based adversarial framework to perform the network alignment tasks. Unlike most existing approaches which formulate the alignment task as the mapping problem between networks, the adversarial learning adopted here is to steer the feature extraction towards alignment tasks by suppressing the domain-dependent features which are considered task-unrelated for network alignment. To best of our knowledge, we are the first to argue that it is helpful to eliminate/suppress the domain-dependent features to improve the performance of network alignment. 
    \item The mathematical models and deductions, and experiments in the paper are specifically tailored to the conventional alignment tasks and tasks involved with directed networks. In particular, the objective function leverages a probabilistic design from a multi-view perspective as the network alignment can be viewed as a bi-directional matching problem. Whereas most of existing approaches adopt an distance-based supervision with the observed anchors.  
    

    \item We evaluate the proposed models with detailed experiments on real-world social network datasets. Results demonstrate significant and robust improvements in comparison with other state-of-the-art approaches.
    
\end{itemize}

The rest of the paper is organized as follows. Section \ref{sec:related-work} summarizes the related work. Section \ref{sec:DANA} illustrates the design and algorithms of vanila \textit{GANA}, and its variations. Section \ref{sec:experiments} reports the experimental design and discusses the results. A case study, which illustrates how the framework suppresses the domain-dependent features to boost the alignment task, is also included in Section \ref{sec:experiments}. Section \ref{sec:conclusion} concludes the paper.

\begin{figure*}[t]
\centering
\includegraphics[width=0.68\textwidth]{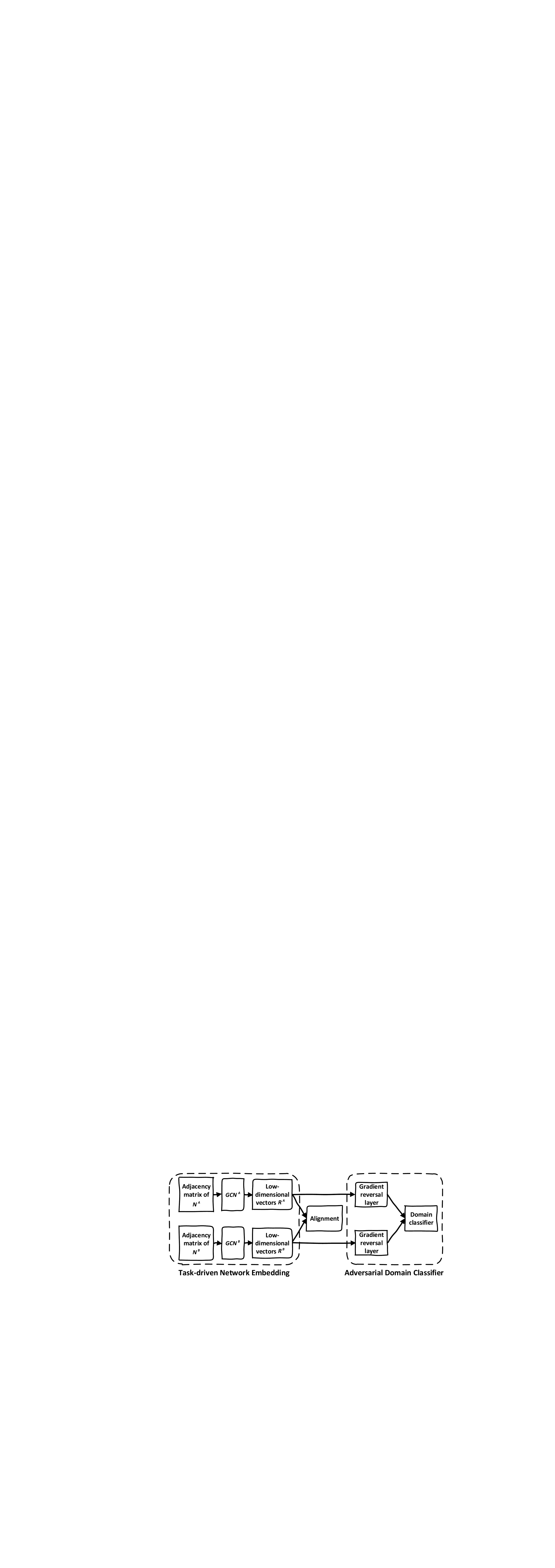}
\caption{\label{fig:model_framework_1}The Vanilla Architecture of DANA}
\end{figure*}
\section{Related Work}\label{sec:related-work}

Our work is most related to embedding-based network alignment and adversarial learning.
\subsection{Embedding-based Network Alignment}
Among the various representation learning-based network alignment approaches, the main difference lies in the way (1) What kind of network embedding approach is leveraged? (2) Whether the multiple networks are projected onto the same low-dimensional subspace? 

\cite{tan2014mapping} proposed a shallow model \textit{MAH} to align the network manifolds by modeling social graphs with hypergraphs. The manifolds of social networks are projected onto a common embedded space, then the user mapping can be inferred by comparing the distances of users in the embedding space. To scale up, \textit{IONE} \cite{liu2016aligning} proposed an embedding approach by only considering the ``second-order proximity'' of local structures to obtain the common low-dimensional subspace of networks, semi-supervised by the observed anchors.

\textit{ULink} \cite{Mu:2016:UIL:2939672.2939849} was proposed to explore the concept of ``Latent User Space'', the objective of which is to find projections of each network while minimizing the distance between the node and its correspondence among their respective vector spaces. Similarly, \textit{PALE} \cite{man2016predict} proposes to embed the networks individually first by leveraging on network embedding approach, e.g., LINE \cite{tang2015line} or Deepwalk \cite{Perozzi2014DeepWalk}, then to seek an explicit feature space transformation that would map one into the other one. However, the standalone embedding process in a two-phase approach like \textit{PALE} is designed irrelevant to the alignment task, thus may not include the features which directly benefit the alignment. And all the aforementioned approaches neglect the importance of learning domain-invariant features.

\subsection{Adversarial Training of Neural Networks}
Generative Adversarial Networks (GANs) \cite{goodfellow2014generative}, which plays an adversarial minimax game between the generator and discriminator, frees the users from the painful practice of defining a tricky objective function. \textit{GANs} shows its impressive potential in various fields/tasks, e.g., natural language processing \cite{zhang2017adversarial,wang2017irgan} and network embedding \cite{dai2018adversarial,wang2018graphgan}. 

Recently, an adversarial training framework \textit{DANN} \cite{ganin2016domain} was proposed for domain adaption. In particular, \textit{DANN} introduces a representation learning module for better domain adaptation, in which the adversarial training pushes maximizing the loss of the domain classifier thus to encourage domain-invariant features to dominate the process of minimizing the loss of the label classifier. \cite{xie2017controllable} further extended this idea to obtain a controllable invariance through adversarial feature learning. Both two approaches were based on the theory that a good representation for domain adaption is one for which an algorithm cannot identify the domain of its input. This is also the building block of our work.

\textit{SNNA} \cite{SNNA} is recently proposed to perform social network alignment via supervised adversarial learning. \textit{SNNA} is a two-phase approach which first learns the low-dimensional representation for each network via the conventional network embedding, then learns the projection function within a \textit{GAN} framework. Supervised by the observed anchors, the generator targets at learning a transformation from one embedding space to another which minimize the Wasserstein distance between the projected source distribution and the target distribution, while the discriminator estimates the distance between two embedding space. In other words, the adversarial learning in \textit{SNNA} is used to obtain an optimal projection function between the two subspaces. 

In contrast to the two-phase \textit{SNNA}, our proposed approach performs network representation learning and alignment learning in a unified architecture. The adversarial learning is mainly for the domain classifier to filter away the domain-dependent feature by maximizing the loss of the classifier. Meanwhile, the presentation learning is also task-driven by maximizing the posterior probability of the observed anchors, thus to produce useful feature representations for network alignment.

\section{Domain-Adversarial Network Alignment}\label{sec:DANA}

In this section, we formulate our problem first, and then present a vanilla framework for domain-adversarial network alignment. Its adaptions with weight-sharing for model simplification and a direction-aware structure for directed networks are further introduced. 

For the same user in different social networks, namely $v^A_i$ in network $A$ and $v^B_j$ in network $B$, we denote $(v^A_i, v^B_j)$ as a pair of anchors. The network alignment task could be formulated as predicting the anchor pair $(v^A_i, v^B_j)$ given two networks $N^A=(V^A, E^A)$ and $N^B=(V^B, E^B)$, where $v^{A/B}_i \in V^{A/B}$, $V^{A/B}$ and $E^{A/B}$ are the sets of vertices and edges in network $A/B$ respectively. Each vertex is either labeled as $d^A$ or $d^B$, indicating the network which the vertex belongs to. Note that we argue that domain-dependent features, which are capable to reveal the domain identity, are futile, sometimes detrimental to alignment task. To achieve better alignment performance, we adopt the domain-adversarial training paradigm to train a domain classifier, which helps to extract domain-invariant representations of networks.


\subsection{Vanilla Architecture of DANA}\label{sec:model-1}

The vanilla architecture of \textit{DANA} consists of two components, namely, task-driven network embedding module and adversarial domain classifier. 

\begin{algorithm}[t]
    \caption{Training procedure of \textit{DANA}}
    \label{alg:model_1}

    \textbf{Input}: network $A$ including $V^A$ and $M^A$, network $B$ including $V^B$ and $M^B$, and the set of anchor seeds $S$.\\
    \textbf{Hyperparameters}: the batch size of vertices $U$; the batch size of anchor seeds $Z$; the weighting factor $\gamma$; the regularization factor $\lambda$.\\
    
    \textbf{Parameters}: the feature extractors ${GCN}^A$: $\Theta_{g^A} = \{\!H^A_0, W^A_l\}$ and ${GCN}^B$: $\Theta_{g^B} =\{H^B_0, W^B_l\}$ where $l = \{1,2,...,L\}$; the domain classifier parameterized as MLP: $\Theta_{\mathcal{D}}$.\\
    
    \textbf{Output}: representations of $V^A$: $R^A=H^A_L$; representations of $V^B$: $R^B=H^B_L$.\\
    
    \begin{algorithmic}[1] 
        \STATE Randomly initialize \{$\Theta_{g^A}, \Theta_{g^B}, \Theta_{\mathcal{D}}\} \sim N(0,I)$
        \REPEAT
            \STATE Sample a batch of vertices from $V^A$: $V^A_U = {\{v^A_u\}}^{U}_{u=1}$\\
            \STATE Sample a batch of vertices from $V^B$: $V^B_U = {\{v^B_u\}}^{U}_{u=1}$\\
            \STATE Sample a batch of anchors from $S$: $S_Z = {\{s_z\}}^{Z}_{z=1}$\\
            \STATE Update $\Theta_{g^A}, \Theta_{g^B}$ with Adam Optimizer to minimize:
                \hspace{8pt}$-\sum_{(v^A_i, v^B_j)\in S_z}\log\frac{1}{2}\left(p(v^B_j|v^A_i) + p(v^A_i|v^B_j)\right)$\\
                \hspace{8pt}$ + \gamma\sum_{v \in \{V^A_U\cup V^B_U\}} \sum_{d\in{\{d^A, d^B\}}} {\mathbb{I}_d(v)}\log p(d|v)$\\
                \hspace{8pt}$+\lambda(\Vert\Theta_{g^A}\Vert + \Vert\Theta_{g^B}\Vert)$
            \STATE Update $\Theta_{\mathcal{D}}$ with Adam Optimizer to minimize:\\
                \hspace{0pt}$-\sum_{v \in \{V^A_U\cup V^B_U\}} \sum_{d\in{\{d^A, d^B\}}} {\mathbb{I}_d(v)}\log p(d|v) +\lambda\Vert\Theta_{\mathcal{D}}\Vert$
        \UNTIL{convergence}
    \end{algorithmic}
\end{algorithm}
\subsubsection{Task-driven Network Embedding} To explore the structural information of networks, we employ \textit{GCNs} as our task-driven feature extractors. Note that we adopted a \textit{GCN} for each network (See Fig.\ref{fig:model_framework_1}). In the following, we omit the superscript $A/B$ which denotes the identity of the network for simplicity. Given the adjacency matrix $M \in \mathbb{R}^{|V|\times |V|}$ of one network, \textit{GCN} outputs the corresponding hidden representations $H_{l} \in \mathbb{R}^{|V|\times k_l}$ in the $l$-th layer with $k_l$ neurons following the layer-wise propagation rule, namely:
\begin{equation}
H_{l} = \sigma(FH_{l-1}W_{l})
\end{equation}
where $l\!=\!\{1,2,...,L\}$. $F\!=\!{D}^{-\frac{1}{2}} (M+I){D}^{-\frac{1}{2}}$ is the convolution kernel, which acts as a spatial filter on network.
$D$ denotes the diagonal node degree matrix of the network, i.e. $D_{ii}={\sum}_jM_{ij}$ and $I$ is the self-connection identity matrix of the network. $W_{l} \in \mathbb{R}^{k_{l-1}\times k_{l}}$ denotes the trainable weight matrix of the $l$-th layer. $H_0$ can be either previously encoded vectors carrying privilege information of the network or randomly initialized. The activation function $\sigma$ is implemented by $ReLU(\cdot)$ in our framework following \cite{kipf2016semi}. Thereby, the \textit{GCN} module outputs a low-dimensional vector $R=H_{L}$ for each network, respectively.
To integrate the representation learning into the alignment task, we optimize the network alignment problem by maximizing the following posterior:
\begin{equation}
\mathcal{P}(\Theta_{g^A}, \Theta_{g^B}|S) \propto \mathcal{P}(S|\Theta_{g^A}, \Theta_{g^B})\mathcal{P}(\Theta_{g^A}, \Theta_{g^B})
\end{equation}
where $S$ denotes the collection of anchor pairs. $\Theta_{g^A}$ denotes all the parameters of the ${GCN}^A$ module, i.e., $\Theta_{g^A}=\{H^A_0, W^A_1, W^A_2,...,W^A_L\}$. The notation definition applies to $\Theta_{g^B}$.
Note that the probability expansions for an anchor pair $(v^A_i,v^B_j)\in S$, i.e.: 
$$\begin{aligned}p(v^B_j,v^A_i| \Theta_{g^A}, \Theta_{g^B}) &= p(v^A_i| \Theta_{g^A}, \Theta_{g^B})p(v^B_j|v^A_i, \Theta_{g^A}, \Theta_{g^B})\\&=p(v^B_j| \Theta_{g^A}, \Theta_{g^B})p(v^A_i|v^B_j, \Theta_{g^A}, \Theta_{g^B}),\end{aligned}$$ are both significant to our problem. We abbreviate $p(v^A_i,v^B_j|\Theta_{g^A}, \Theta_{g^B})$ to $p(v^A_i,v^B_j)$, then we have  $p(v^B_j|v^A_i)$ and $p(v^A_i|v^B_j)$ as the abbreviations of $p(v^B_j|v^A_i, \Theta_{g^A}, \Theta_{g^B})$ and $p(v^A_i|v^B_j, \Theta_{g^A}, \Theta_{g^B})$, respectively. Therefore, we define $p(v^B_j,v^A_i) = 1/2 (p(v^A_i)p(v^B_j|v^A_i) + p(v^B_j)p(v^A_i|v^B_j))$, which is a popular practice for multi-view problems where all views matter. Further, a Gaussian prior is introduced for the model parameters, i.e. $p(\Theta_{g^A}) \sim N(0,I)$ and $p(\Theta_{g^B}) \sim N(0,I)$. The resultant optimization criterion $\mathcal{J}_e$ can be derived as follows:
\begin{equation}
 \label{fun:loss_g1}
 \resizebox{.92\linewidth}{!}{$
    \displaystyle
\mathcal{J}_e=\mkern-11mu\sum_{(v^A_i, v^B_j)\in S}\mkern-11mu\log\frac{1}{2}\mkern-4mu\left(p(v^B_j|v^A_i)p(v^A_i)+p(v^A_i|v^B_j)p(v^B_j)\right)-\lambda(\Vert\Theta_{g^A}\Vert+\Vert\Theta_{g^B}\Vert)$}
\end{equation}
where $p(v^A_i)$ and $p(v^B_j)$ are the constants.
Softmax function is used to approximate the likelihood of observing an anchor pair, namely:
\begin{subequations}
\begin{align}
\label{fun:p_a}
p(v^B_j|v^A_i)&=\frac{\exp(r^B_j\cdot r^A_i)}{\sum_{n=1}^{|V^B|} \exp(r^B_n\cdot r^A_i)}\\
p(v^A_i|v^B_j)&=\frac{\exp(r^A_i\cdot r^B_j)}{\sum_{n=1}^{|V^A|} \exp(r^A_n\cdot r^B_j)}
\label{fun:p_b}
\end{align}
\end{subequations}
where $r_i^A$ corresponds to the learned representation of vertex $v^{A}_i \in V^{A}$. The same is true for $r_i^B$. Due to the summation over the entire set of nodes in Eq.\eqref{fun:p_a} and Eq.\eqref{fun:p_b}, it will be time-consuming for large scale networks. To reduce the
computational complexity, we adopted a sampled softmax function \cite{jean2014using}, which performs the summations over a set of sampled candidates, namely
\begin{equation}
p(v^B_j|v^A_i)=\frac{\exp(r^B_j\cdot r^A_i)}{\sum_{v_c \sim \mathcal{P}^B(v)}^{|C^B|} \exp(r^B_c\cdot r^A_i)}.
\end{equation}
The candidate set
$C^B\!\subset\!V^B$ is sampled via a log-uniform distribution $\mathcal{P}^B(v)$. Such operation also applies to Eq.\eqref{fun:p_b}.

\subsubsection{Adversarial Domain Classifier} However, the optimization criterion Eq.\eqref{fun:loss_g1} could not induce purging the task-irrelevant domain feature, which may weaken the professionalism of representations for network alignment. Inspired by the adversarial learning paradigm, we further augment the alignment task-driven network embedding with an adversarial learning to a domain classifier, which is meant to filter away the domain-dependent features while concentrating on extracting alignment-targeted features.


Note that the domain classifier, acting as the discriminator, tries to distinguish which domain a given vertex $v \in \{V^A\cup V^B\}$ comes from, while feature extractors, i.e. GCNs in our framework, act as a role of the generator, aiming at learning domain-invariant features from the input data to fool the domain classifier. Technically, the domain classifier and the feature extractor are trained by playing minimax games expressed as follows:
\begin{equation}
\underset{\Theta_{g^A},\Theta_{g^B}}{\max}\underset{\Theta_{\mathcal{D}}}{\min} \mathcal{J}_d = \sum_v  \sum_d -{\mathbb{I}_d(v)}\log p(d|v)
\end{equation}
where $d \in \{d^A, d^B\}$ denotes the label of the domain $v$ belongs to, and $\Theta_{\mathcal{D}}$ is the parameter set of the domain classifier. Note that $\mathbb{I}_d(v)$ is the indicator function, which equals to 1 if $v$ comes from the domain $d$ and 0 otherwise. We employ an MLP classifier where the last hidden layer is connected to a softmax layer to induce the conditional distribution $p(d|v)$.

Referring back to Eq.\eqref{fun:loss_g1} for the network alignment task, we train ${GCN}^A$ and ${GCN}^B$ to extract domain-invariant feature representations while maximizing the posterior probability for network alignment with the following form:
\begin{equation}\label{overallobj}
\max_{\Theta_{g^A},\Theta_{g^B}}\min_{\Theta_{\mathcal{D}}}\mathcal{J} = \mathcal{J}_e + \gamma\mathcal{J}_d
\end{equation}
where hyperparameter $\gamma$ is a weighting factor to modulate the contribution of $\mathcal{J}_d$. To optimize $\Theta_{g^A}$, $\Theta_{g^B}$ and $\Theta_{\mathcal{D}}$, we incorporate a Gradient Reversal Layer (GRL) \cite{ganin2016domain} between feature extractors and domain classifier. GRL can be viewed as an activation function layer with no parameters, which identically transfers the input during the forward pass but reverses gradients (multiplied by $-1$) during the back propagation. The adoption of GRL enables a synchronous optimization of Eq.\eqref{overallobj}, thus  \textit{DANA} can be trained easier and faster. The overall architecture and algorithm of our proposed model are depicted in Fig.\ref{fig:model_framework_1} and Algorithm \ref{alg:model_1}, respectively.

\begin{figure}[t]
\centering
\includegraphics[width=0.49\textwidth]{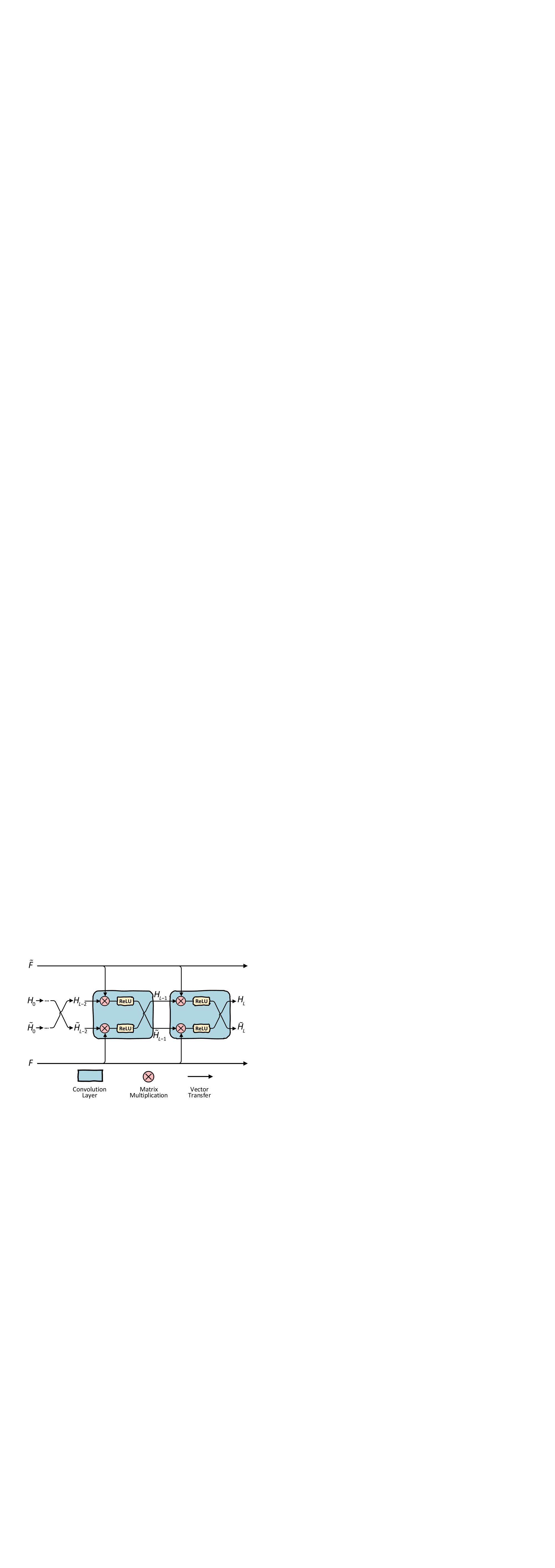}
\caption{\label{fig:Unfoldstructures}Unfolded structure for directed networks}

\end{figure}
\subsection{DANA for Directed Networks}\label{sec:directed-gcn}
There exist many networks deliberately defined as the directed graph. For example, Twitter created a directed graph of followers because the interactions in Twitter are generally one-way. Stemmed from the spectral graph theory, the conventional GCN requires a symmetric adjacency matrix to obtain the low-dimensional representation, which makes our model limited to dealing with the undirected graph. To address directed networks, existing research simply relaxes the strict constraint on the symmetric adjacency matrix in GCNs, and explains the convolutional kernel from a spatial perspective \cite{schlichtkrull2018modeling}. However, it suffers an inadequate characterization of the directed edges in networks, which is important for obtaining accurate representations of the associated vertices. In pursuit of better representations, we elaboratively characterize each vertex from two perspectives, which performs the convolution according to its in-degree and out-degree distributions, respectively.



Given an adjacency matrix $M$ of a directed network, and randomly initialized $H_0$ and $\widetilde{H}_0$, the hidden representation of $H_l$ and $\widetilde{H}_l$ in the $l$-th layer can be obtained as follows:
\begin{subequations}
\begin{align}
H_l &= \sigma(F \widetilde{H}_{l-1}W_l)\label{outd}\\
\widetilde{H}_l &= \sigma(\widetilde{F} H_{l-1}\widetilde{W}_l)\label{ind}
\end{align}
\end{subequations}
where $F^A={D}^{-1}(M+I)$, ${\widetilde{F}}$ = $\widetilde{D}^{-1}(\widetilde{M}+I)$, and $\widetilde{M} = {M}^T$, $\widetilde{D}_{ii}={\sum}_j\widetilde{M}_{ij}$. Eq.\eqref{outd} focuses on the convolution operations on vertices' out-going neighbours, and Eq.\eqref{ind} focuses on the convolution operations on vertices' in-going neighbours. At length, each {GCN} outputs two low-dimensional representations for each vertex, i.e. $R=H_L$ and $\widetilde{R}=\widetilde{H}_L$. The computation and dataflow through the unfolded structure are also depicted in Fig.\ref{fig:Unfoldstructures}. Then, $r_i$ and $\widetilde{r}_i$ of each vertex $v_i$ are concatenated to perform the alignment.

\begin{figure*}[t]
    \subfigure[DBLP]{
    \includegraphics[width=0.365\textwidth]{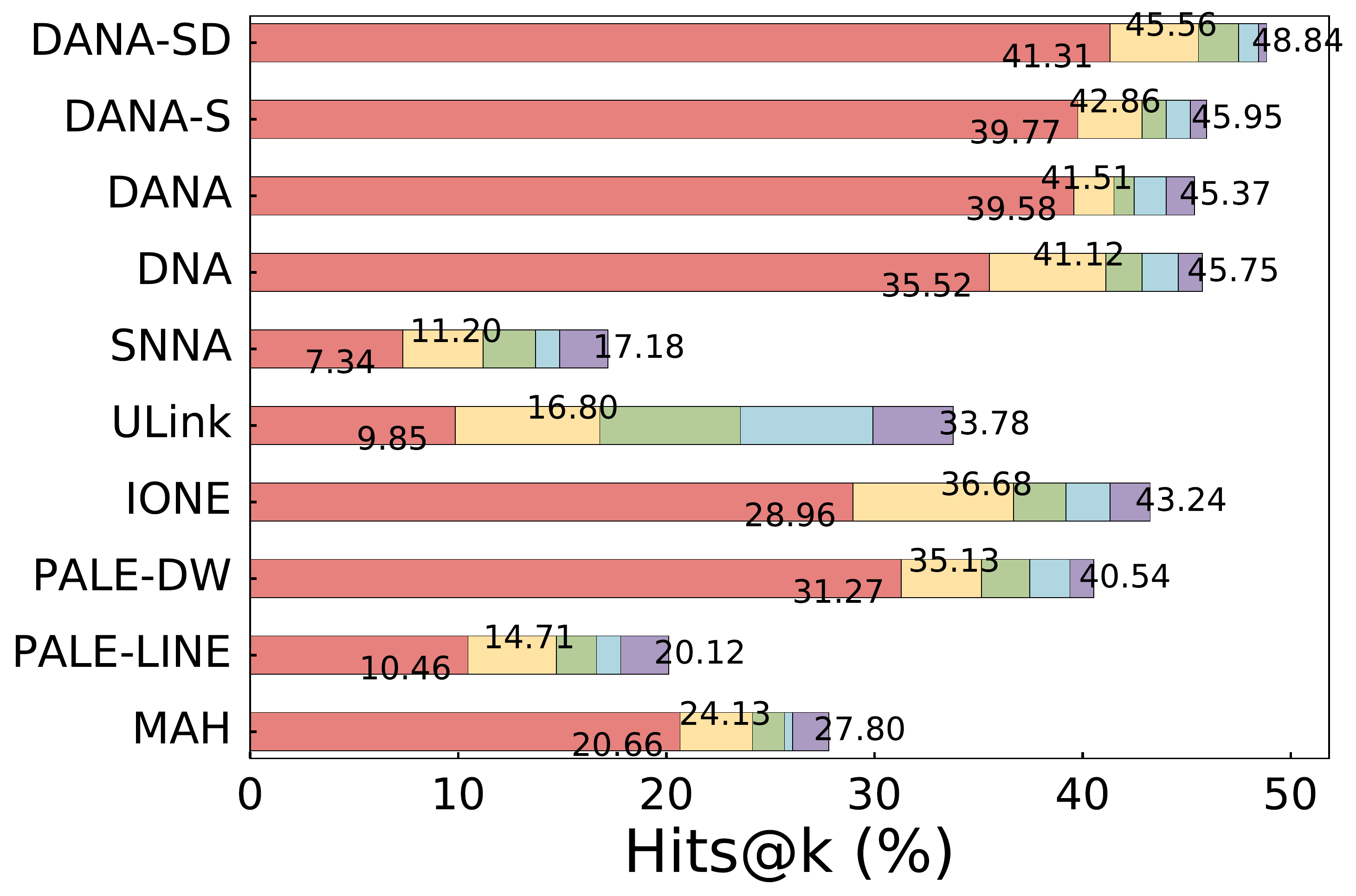}
    \label{fig:perform-dblp}
    }
    \subfigure[Foursquare-Twitter]{
    \includegraphics[width=0.311\textwidth]{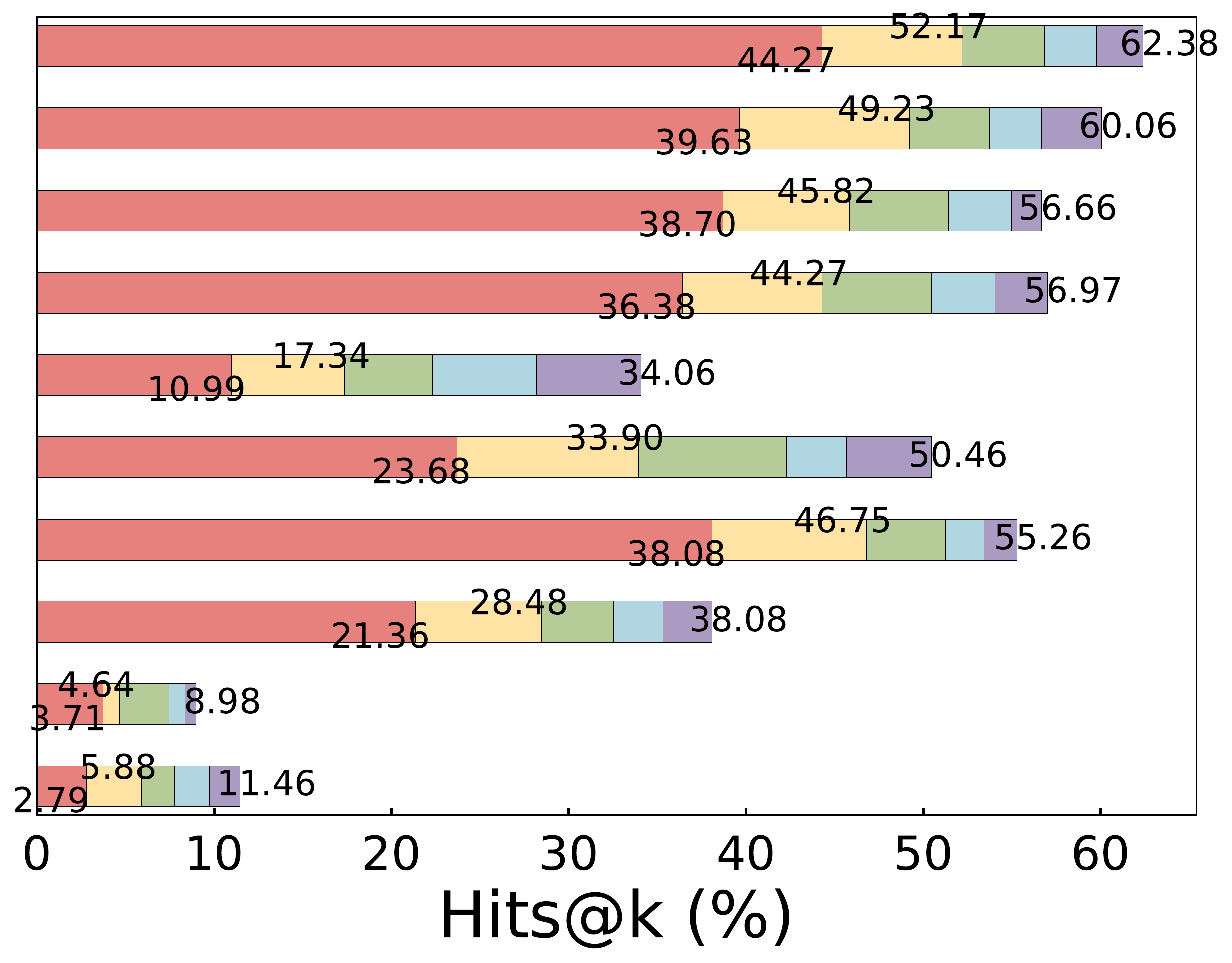}
    \label{fig:perform-ASN-FoT}
    }
    \subfigure[Douban-Weibo]{
    \includegraphics[width=0.317\textwidth]{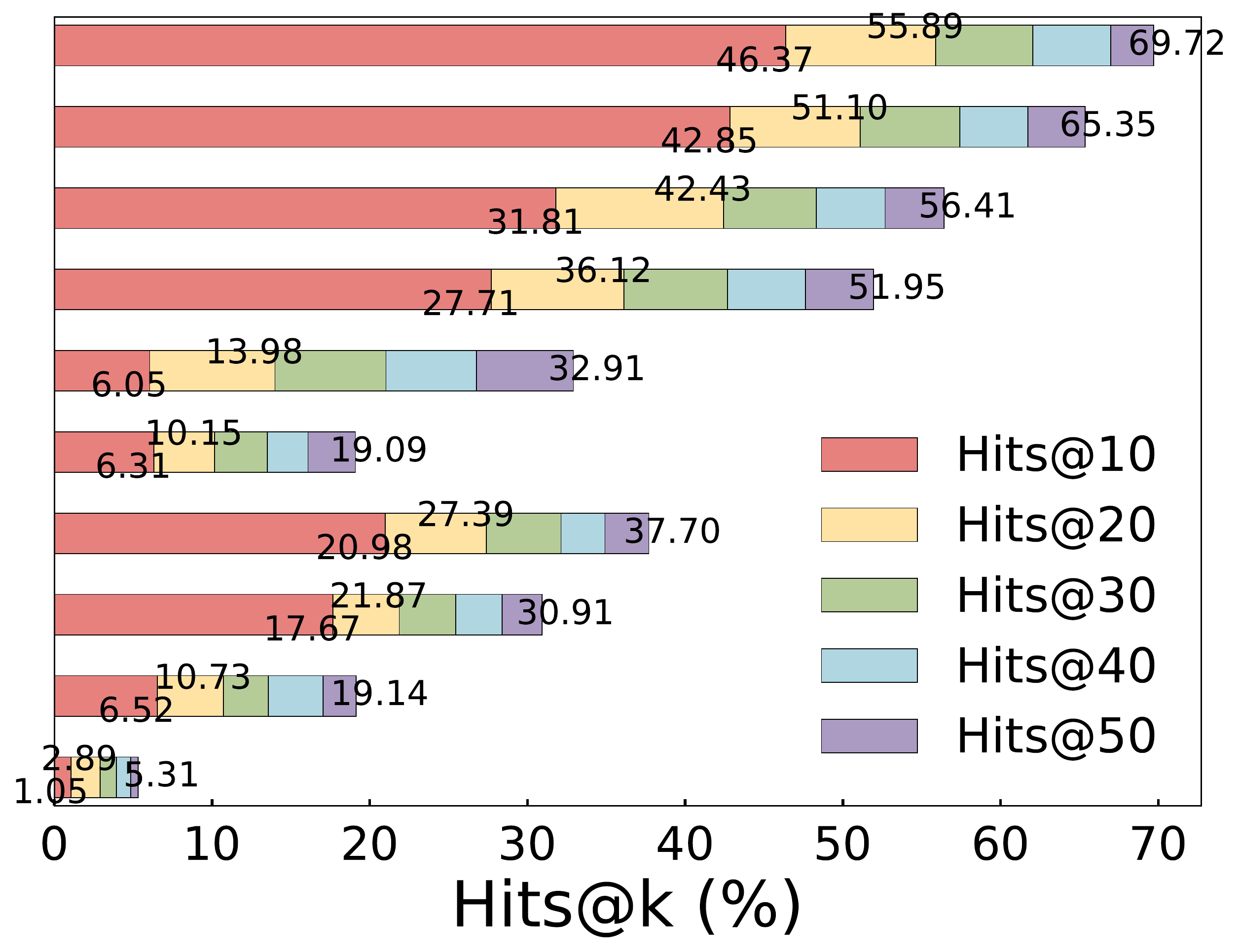}
    \label{fig:perform-ASN-DW}
     }
\caption{Detailed performance comparison on real-world datasets.}
\label{fig:perform-comparison}
\end{figure*}

\subsection{Weight-sharing Between GCNs}\label{sec:domain-shared-weights}
An ideal representation learning for alignment task is to obtain a low-dimensional subspace in which the two vertices of an anchor pair are close to each other. Thus the candidates of a vertex can be obtained based on a ``distance'' between the two vectors. Drawing the subspaces close to each other is usually supervised by forcing the vertices of an anchor pair to share the same representation.

In this paper, we further reinforce the closeness between subspaces by sharing weights across the two GCNs i.e. enforcing $W^A_l = W^B_l, l=\{1,2,...,L\}$. Additionally, such weight-sharing reduces the number of parameters and simplifies our model so that it is more favorable to model training.

\section{Experiments}\label{sec:experiments}
In this section, we present the experimental evaluations of our proposed models and the competing baselines over three real-world datasets. 

\subsection{Metrics, Datasets and Comparative Models}
\subsubsection{Metrics} We evaluate the performance of our proposed models and competing baselines using a metric of Hits@k:
\begin{equation}
Hits@k = \frac{{Hits}^{A}@k + {Hits}^{B}@k}{ |S_{test}|\times 2}\nonumber
\end{equation}
where ${Hits}^{B/A}@k$ means the number of hits in test set $S_{test}$ given the top-k candidates in network $A/B$ for each vertex from network $B/A$. In our models, the Cosine similarity is adopted as the scoring criteria to obtain the top-k candidate list. For the baselines, the candidate lists are obtained following the scoring criteria suggested in their papers. In addition to hits@k, we also adopted the Mean Reciprocal Rank (MRR) \cite{radev2002evaluating} to evaluate the models. Similar to the definition of ${Hits}@k$, MRR in this paper is an average value of bi-directional counts. 


\subsubsection{Datasets} We employ three real-world cross network data sets, the statistics of which are tabulated in Table \ref{tab:dataset}. For the DBLP \cite{tang2008arnetminer}
dataset, authors are split into two different co-author networks (Data Mining and Machine Learning) by filtering publication venues of their papers. The ground truth anchors of this dataset are the authors who published papers in both areas. Note that the co-author relationships are non-directional in DBLP. In contrast, the other two datasets \cite{zhang2015integrated}\cite{cao2016asnets} are constructed from the directed social networks. The ground truth of the anchor users is obtained based on the fact that some users provide their unified accounts across social networks.

\begin{table}[t]
\centering
\caption{\label{tab:dataset}Statistics of the datasets used for evaluation}
\scalebox{1}{
\begin{tabular}{lrc}  
\toprule
Dataset & Network(\#Nodes, \#Edges) & \#Anchors\\
\midrule
\multirow{2}{*}{DBLP}&Data Mining (11526, 28565) & \multirow{2}{*}{1295}\\
&Machine Learning (12311, 26162) &\\
\midrule
\multirow{2}{*}{Fq.-Tw.}&Foursquare (5313, 76972) & \multirow{2}{*}{1611}\\
&Twitter (5120, 164920)&\\
\midrule
\multirow{2}{*}{Db.-Wb.}&Douban (10103, 527980) & \multirow{2}{*}{4752}\\
&Weibo (9576, 270780) &\\
\bottomrule
\end{tabular}}
\end{table}
\subsubsection{Comparative Models} Our proposed model DANA with its variants and the state-of-the-art baseline methods for comparison are listed as following: \begin{itemize}
\item \textbf{\textit{MAH}} \cite{tan2014mapping}: A hypergraph-based manifold matching approach for network alignment, where the hyperedges model the high-order relations in social networks. 
\item \textbf{\textit{ULink}} \cite{Mu:2016:UIL:2939672.2939849}: An approach for multi-platform user identity linkage predication in which Latent User Space was proposed and utilized. The constrained concave-convex procedure is also adopted for the model inference. 
\item \textbf{\textit{IONE}} \cite{liu2016aligning}: The state-of-the-art approach for network alignment which incorporates the learning of the second-order proximity preserving embeddings and the network alignment in a unified framework. 
\item \textbf{\textit{PALE-LINE}} \cite{man2016predict}: An embedding-based approach where the embeddings of individual networks are learned using LINE~\cite{tang2015line}, and an MLP is used for learning the project function between the low-dimensional subspaces of networks. 
\item \textbf{\textit{PALE-Deepwalk}} \cite{man2016predict}:  A variant of \textit{PALE-LINE}, in which DeepWalk~\cite{Perozzi2014DeepWalk} is adopted for learning individual network embeddings. The projection function learning is the same as that of \textit{PALE-LINE}. 
\item \textbf{\textit{SNNA}} \cite{SNNA}: An adversarial approach to network alignment where the low-dimensional subspaces of networks are obtained by using existing network embedding approaches. The generator is then designed to learn a projection function from one subspace to another, and the discriminator is to estimate the wasserstein distance between the projected source distribution and the target distribution.
\item \textbf{\textit{DANA}}: The vanilla version of our proposed framework in this paper.
\item \textbf{\textit{DANA-S}}: A variation of \textit{DANA} where the Suffix ``-S'' of the name indicates an incorporation with weight-sharing adopted in the model. 
\item \textbf{\textit{DANA-SD} }: A variation of \textit{DANA} where ``D'' further indicates an incorporation of the direction-aware structure on top of \textit{DANA-S}. 
\item \textbf{\textit{DNA} }: refers to a variation of \textit{DANA} where the domain adversarial component (Gradient reversal layer and domain classifier) is removed. 
\end{itemize}

\begin{table*}[ht]
    \centering
    \caption{Hits@1 and MRR comparison on real-world datasets.}
    \begin{tabular}{ccrrrrrrrrrc}
    \toprule
Dataset& Metric & MAH&	PALE-LINE&	PALE-DW&	IONE&	Ulink&	SNNA&	DNA&	DANA&	DANA-S&	DANA-SD\\
\midrule
\multirow{4}{*}{DBLP}&	Hits@1&	0.0695&	0.0277&	0.0772&	0.0560&	0.0116&	0.0096&	0.2104&	0.2182&	0.2201&	\multirow{2}{*}{\textbf{0.2297}}\\
&	Imp(\%)&	230.50&	729.24&	197.54&	310.18&	1880.17&	2292.71&	9.17&	5.27&	4.36&	\\
\cline{2-12}
&	MRR&	0.1108&	0.0422&	0.1710&	0.1414&	0.0503&	0.0312&	0.2739&	0.2830&	0.2838&	\multirow{2}{*}{\textbf{0.2895}}\\
&	Imp(\%)&	161.28&	586.02&	69.30&	104.74&	475.55&	827.88&	5.70&	2.30&	2.01&	\\
\midrule
\multirow{4}{*}{Fq.-Tw.}&	Hits@1&	0.0062&	0.0093&	0.0464&	0.1409&	0.0495&	0.0372&	0.1207&	0.1486&	0.1548&	\multirow{2}{*}{\textbf{0.1842}}\\
&	Imp(\%)&	2870.97&	1880.65&	296.98&	30.73&	272.12&	395.16&	52.61&	23.96&	18.99&	\\
\cline{2-12}
&	MRR&	0.0176&	0.0164&	0.0928&	0.2132&	0.1479&	0.0550&	0.2017&	0.2258&	0.2391&	\multirow{2}{*}{\textbf{0.2579}}\\
&	Imp(\%)&	1365.34&	1472.56&	177.91&	20.97&	74.37&	368.91&	27.86&	14.22&	7.86&	\\
\midrule								

\multirow{4}{*}{Db.-Wb.}&	Hits@1&	0.0032&	0.0126&	0.0358&	0.0794&	0.0074&	0.0042&	0.0847&	0.1420&	0.1772&	\multirow{2}{*}{\textbf{0.1930}}\\
&	Imp(\%)&	5931.25&	1431.75&	439.11&	143.07&	2508.11&	4495.24&	127.86&	35.92&	8.92&	\\
\cline{2-12}
&	MRR&	0.0081&	0.0317&	0.0822&	0.1224&	0.0301&	0.0300&	0.1598&	0.2144&	0.2228&	\multirow{2}{*}{\textbf{0.2608}}\\
&	Imp(\%)&	3119.75&	722.71&	217.27&	113.07&	766.45&	769.33&	63.20&	21.64&	17.06&	\\
	\bottomrule
    \end{tabular}
    \label{tab:alignment_Hits@1_MRR}
\end{table*}

In our experiments, for \textit{DANA} and its variants, we use 2-layer GCNs for feature extractor and a 2-layer MLP for domain classifier. The batch size of vertices $U$ for domain-adversarial training is set to 512 and the batch size of anchors seeds $Z$ is set as the size of the training set. The parameters are optimized using Adam optimizer with a learning rate of 0.001, a weighting factor $\gamma=1.0$, and $\lambda=0.01$ for regularization.
The state-of-the-art approaches, including \textit{MAH} \cite{tan2014mapping}, \textit{ULink} \cite{Mu:2016:UIL:2939672.2939849}, \textit{IONE} \cite{liu2016aligning}, 
\textit{PALE-LINE}, \textit{PALE-Deepwalk} \cite{man2016predict}, and \textit{SNNA} \cite{SNNA}, are evaluated as the competing baselines. They are trained based on the settings recommended in the published papers or the distributed open source code until convergence.

\subsection{Experimental Results}
\subsubsection{Overall Alignment Performance.}
\begin{figure*}[t]
\centering
\begin{minipage}{0.48\textwidth}
    \centering
    \includegraphics[width=0.70\textwidth]{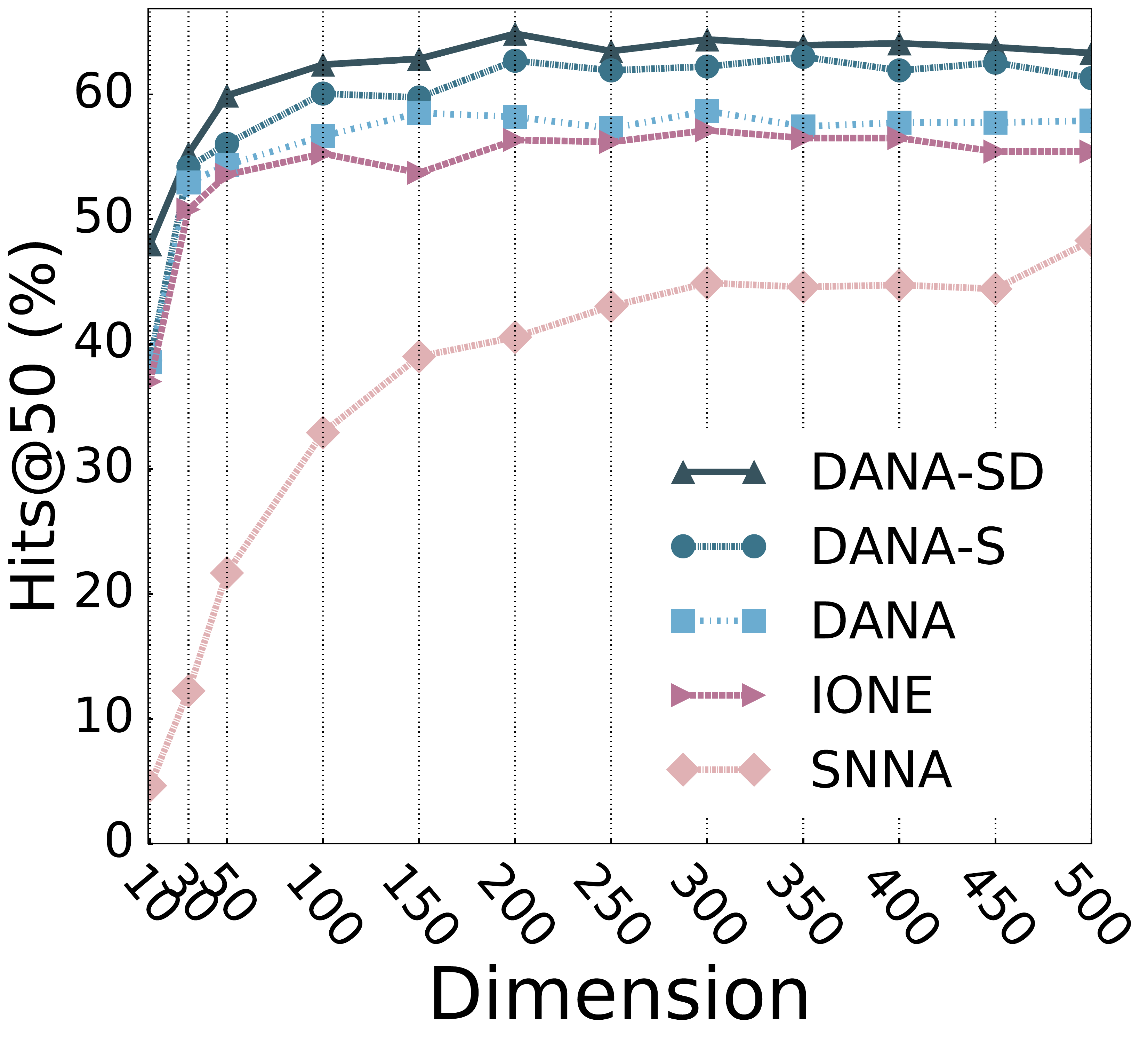}
    \caption{Hits@50 vs. Dimension on Foursquare-Twitter.}
    \label{fig:dimensions}
\end{minipage}
\begin{minipage}{0.48\textwidth}
    \centering
    \includegraphics[width=0.70\textwidth]{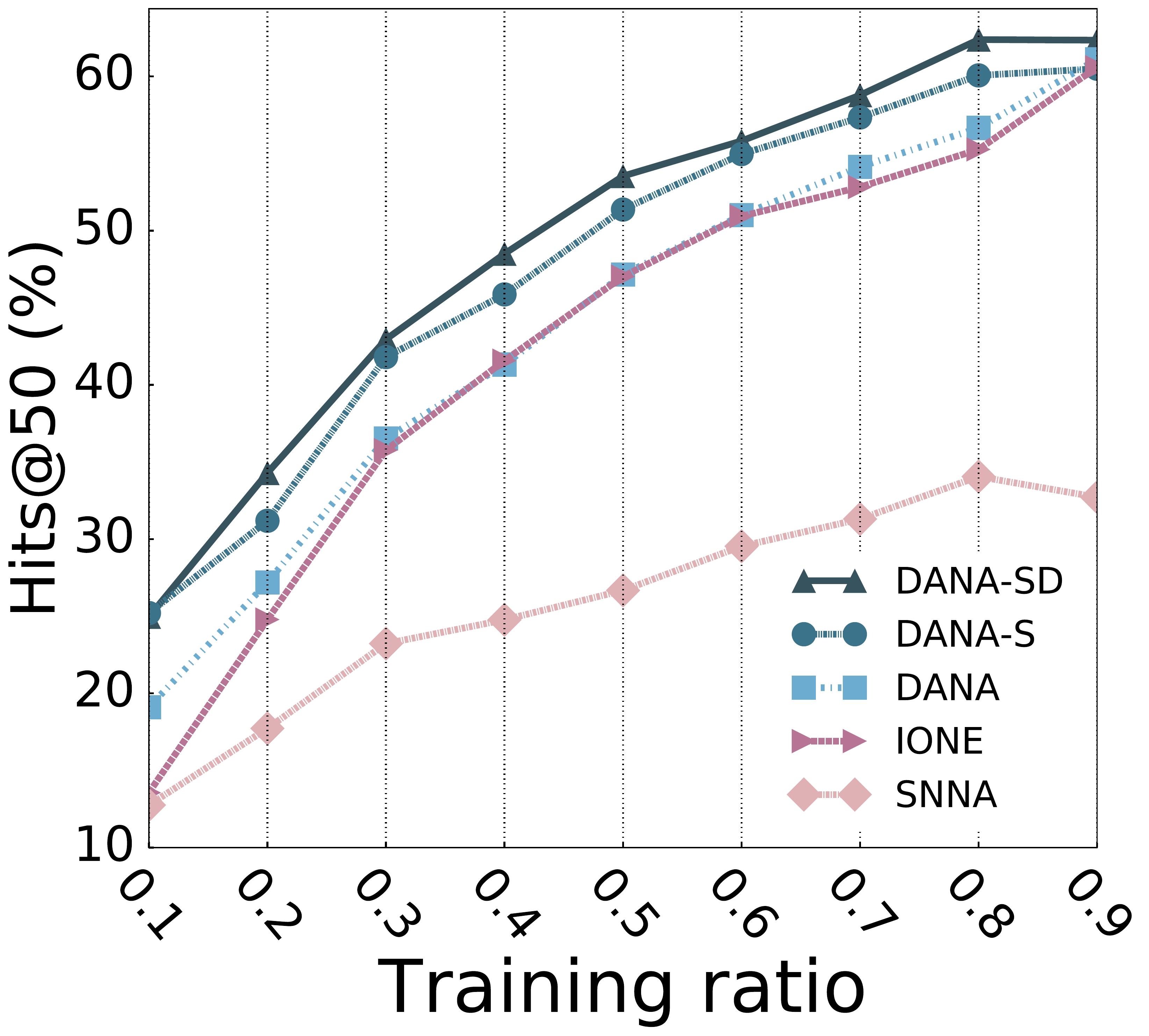}
    \caption{Hits@50 vs. Training ratio on Foursquare-Twitter.}
    \label{fig:trainingratio}
\end{minipage}
\end{figure*}
In this section, we compare the performance of \textit{DANA} with its variations and other baselines on three real-world datasets. We set 80\% of the anchors as the training set and the rest as the test set. The dimension of the embedding is unanimously set to 100 for all models. Note that $k_L$ is set to 50 in \textit{DANA-SD} as the embedding is the concatenation of two vertex representations $r_i$ and $\widetilde{r_i}$. We tabulate Hits@1, MRR and \textit{DANA-SD}'s improvement over all comparative approaches in Table \ref{tab:alignment_Hits@1_MRR}. And the experimental results of Hits@k ($k=\{10,20,30,40,50\}$) are presented in Fig.\ref{fig:perform-comparison}.

From Fig.\ref{fig:perform-comparison} and Table \ref{tab:alignment_Hits@1_MRR}, we can observe that:
\begin{enumerate}
    \item \textit{DANA} and its variants significantly outperform most baselines, under different @K settings for all datasets. It demonstrates the efficacy of the proposed \textit{DANA} framework. In particular, \textit{DANAs} improve Hits@1 by 190+\%, 30+\% and 140+\% respectively over the most competitive baseline on DBLP, Foursquare-Twitter and Douban-Weibo. When $k$ becomes larger, \textit{DANAs} can still achieve more than 15+\% performance improvement. In general, the improvement becomes more significant when k is smaller.

    
    \item The unified frameworks, e.g., \textit{IONE}, achieve much higher accuracy than the two-phase methods, e.g, \textit{PALE-LINE} and \textit{PALE-Deepwalk}. Because the embedding process (first-phase) in two-phase framework is independent of the objective of the alignment task, which would result in unsuitable representations for the transformation process in the second-phase. Besides, the two-phase alignment method is also sensitive to the adopted embedding approach (e.g., \textit{Deepwalk} performs better than \textit{LINE} in \textit{PALE} framework).
    \item Both \textit{ULink} and \textit{SNNA} do not perform well with only the structural information, as they heavily rely on the initialization of the embedding. In particular, better performances of \textit{ULink} and \textit{SNNA} usually come with the initialization using the privilege information, e.g., attributes. Whereas, benefiting from the adopted GCNs, \textit{DANA} and its variants are robust to the initialization. 
    \item The matrix factorization-based approach \textit{MAH} performs worst because matrix-factorization is kind of linear method which is usually inferior to the non-linear embedding method used in our framework. Further, \textit{MAH} is hard to scale up for large-scale problems due to the matrix inversion involved. For Foursquare-Twitter dataset, \textit{MAH} requires the representation with over 800 dimensions to reach convergence \cite{liu2016aligning}, which further validates the efficiency of the embedding-based approaches.
\end{enumerate}

Compared with \textit{DANA} and its variants, \textit{DNA} (\textit{DANA} without the adversarial learning module) achieves lower accuracy. It demonstrates the effectiveness of the domain adversarial learning w.r.t. the network alignment task. Benefiting from the introduced weight-sharing structure, \textit{DANA-S} performs better than the vanilla \textit{DANA}. \textit{DANA-SD} outperforms all the baselines which validates the importance of the incorporation of direction-aware structure. Note that \textit{DANA-SD} also achieves a performance enhancement on the undirected network DLBP, we believe it's due to the larger parameter set (an adoption of $\widetilde{W}$). The superiority of \textit{DANA-SD} becomes more obvious for larger directed networks, i.e. Douban-Weibo dataset. We also investigate the importance of directional edges to the entire network via analyzing network structures. It turns out that the number of connected components and that of strongly connected components in Foursquare-Twitter differ significantly compared with Douban-Weibo dataset. It indicates the direction information play a rather important role in the Foursquare-Twitter dataset. Thus, Foursquare-Twitter dataset may be beneficial to the LINE-based model \textit{IONE} which joints three sets of vectors from different views for directed network alignment \cite{liu2016aligning}. In comparison, \textit{DANA-SD} employs two sets of vectors to capture the directions, but still improves Hits@k by 10\%+ over \textit{IONE}.

Fig.\ref{fig:dimensions} and Fig.\ref{fig:trainingratio} show the outperformance of \textit{DANA-SD} on the Foursquare-Twitter dataset, given different dimension settings as well as different training-to-test ratios. Fig.\ref{fig:trainingratio} also indicates that, in a weakly-supervised manner, our proposed models can still achieve robust and obvious outperformance. 


To sum up, we have \textit{DANA-SD}$>$\textit{DANA-S}$>$\textit{DANA}$>$\textit{DNA} in terms of alignment accuracy, which is consistent with our motivation in this paper.

Regarding the efficiency, \textit{DANA} and its variants take few minutes (within 500 epochs) to reach convergence, which is much faster compared with other baselines. That is because: (1) GCNs is an efficient feature extractor. (2) the gradient reversal layer enables synchronous learning of Eq.\eqref{overallobj}.

\begin{figure}[t]
\centering
    \subfigure[DBLP]{
        \includegraphics[width=0.243\textwidth]{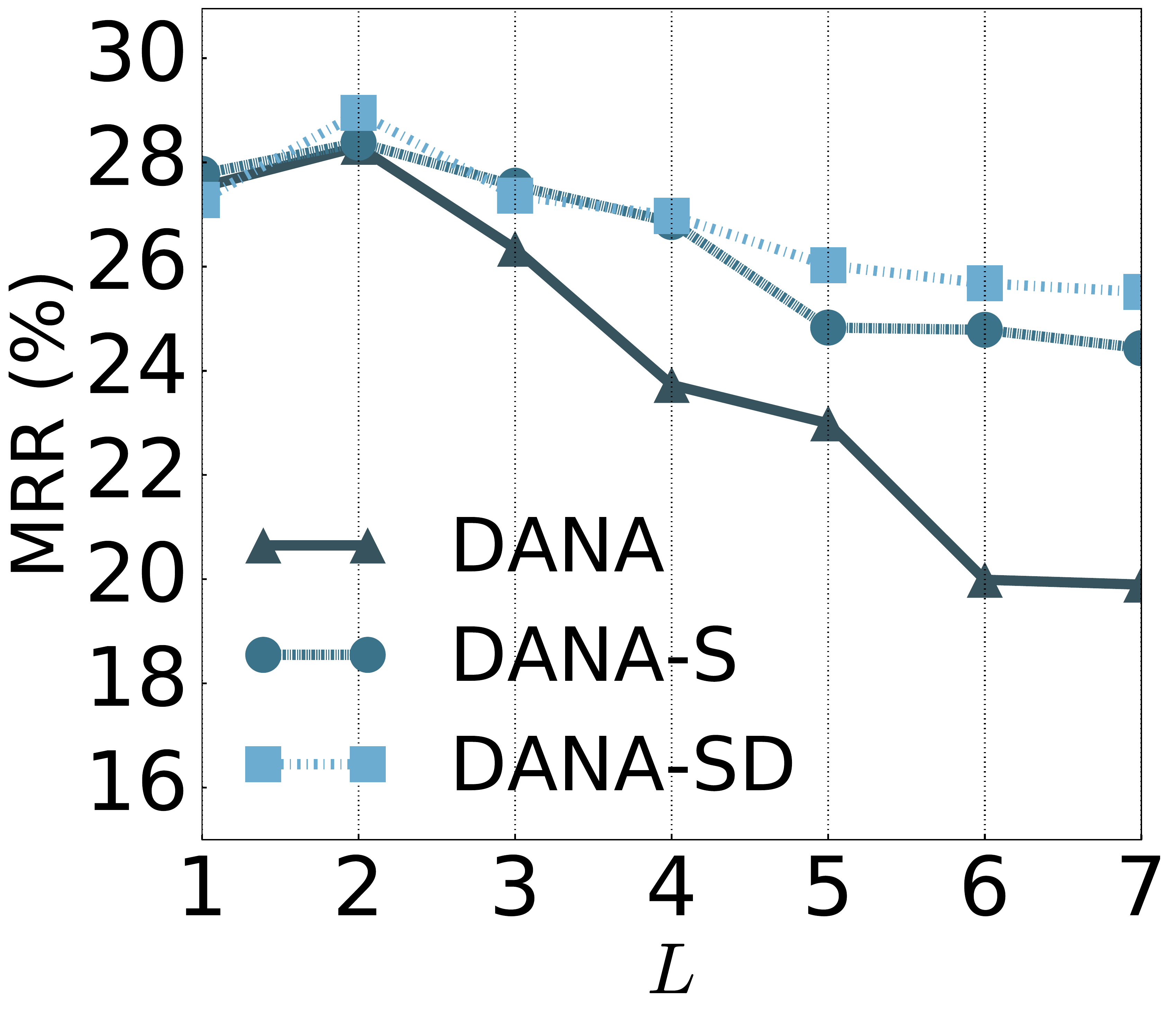}
        \label{fig:layer_dblp}
    }\subfigure[Foursquare-Twitter]{
        \includegraphics[width=0.243\textwidth]{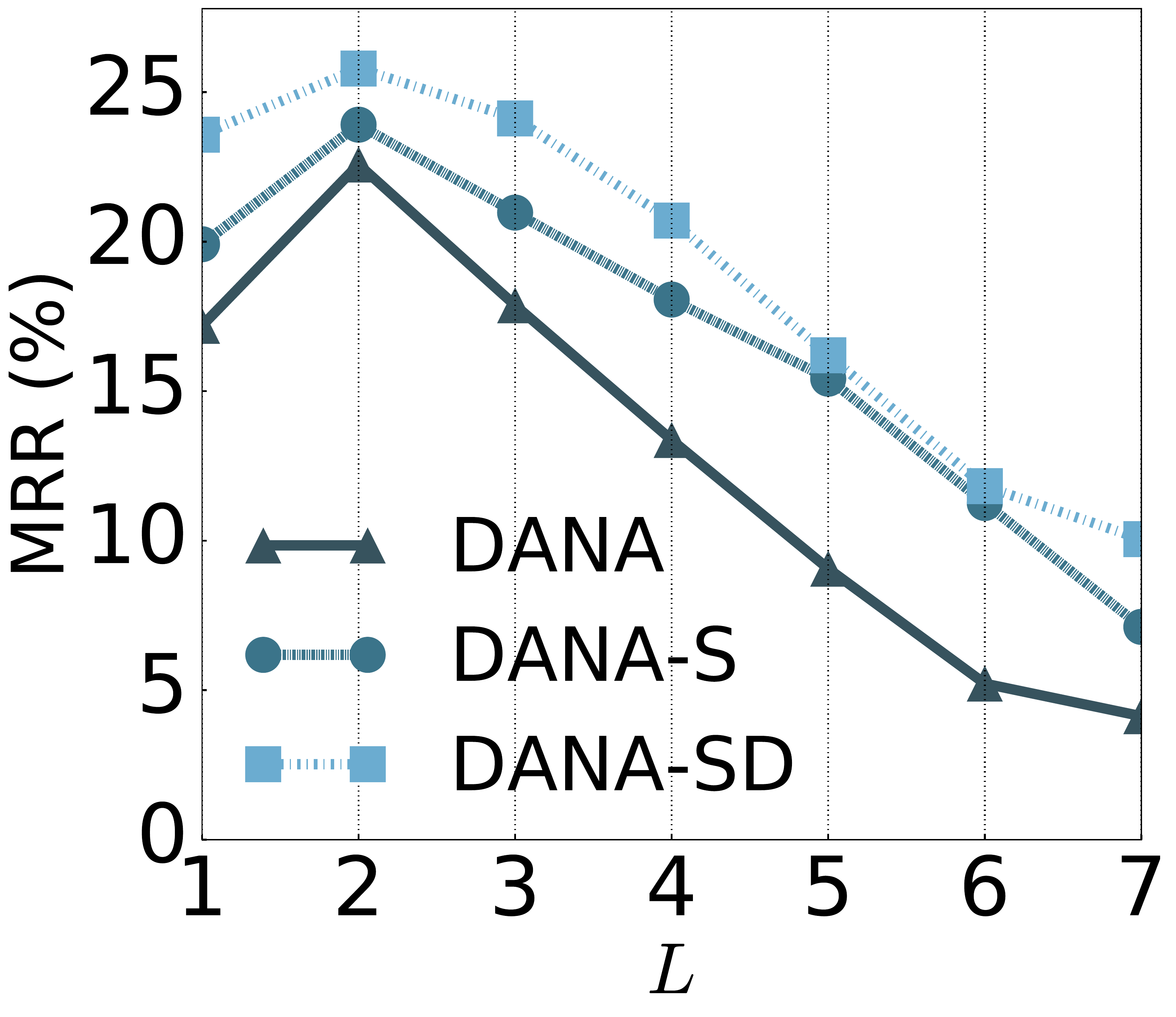}
        \label{fig:layer_fo-tw}
    }
\caption{\label{fig:layers} Sensitivity analysis of parameter $L$}
\end{figure}
\subsubsection{Parameter Sensitivity Analysis}
To analyze the effects of the hyperparameters in \textit{DANAs} which are the number of layers in GCNs $L$ and the weighting factor $\gamma$, we conduct the experiments of \textit{DANAs} with different L-layers GCN and different values of $\gamma$.

In Fig.\ref{fig:layers}, we vary the number of the layers (from 1 to 7) in GCNs, as well as fixing all other parameters. And we observe that DANAs achieve the best performance with the 2-layers GCNs. When $L>2$, the deeper layers GCNs have, the worse the performance. The observation is consistent with the general acknowledgement that two-layers usually are the best setting for the conventional GCNs \cite{DBLP:conf/aaai/LiHW18}. That is because the graph convolution of the GCN model can be viewed as a special form of Laplacian smoothing over the features of a vertex and its nearby neighbors. However, the operation also results in an over-smoothing when involved with many convolutional layers, leading the output features of vertices less distinguishable and an inferior alignment performance.




\begin{figure}[t]
    \subfigure[Foursquare-Twitter]{
        \includegraphics[width=0.243\textwidth]{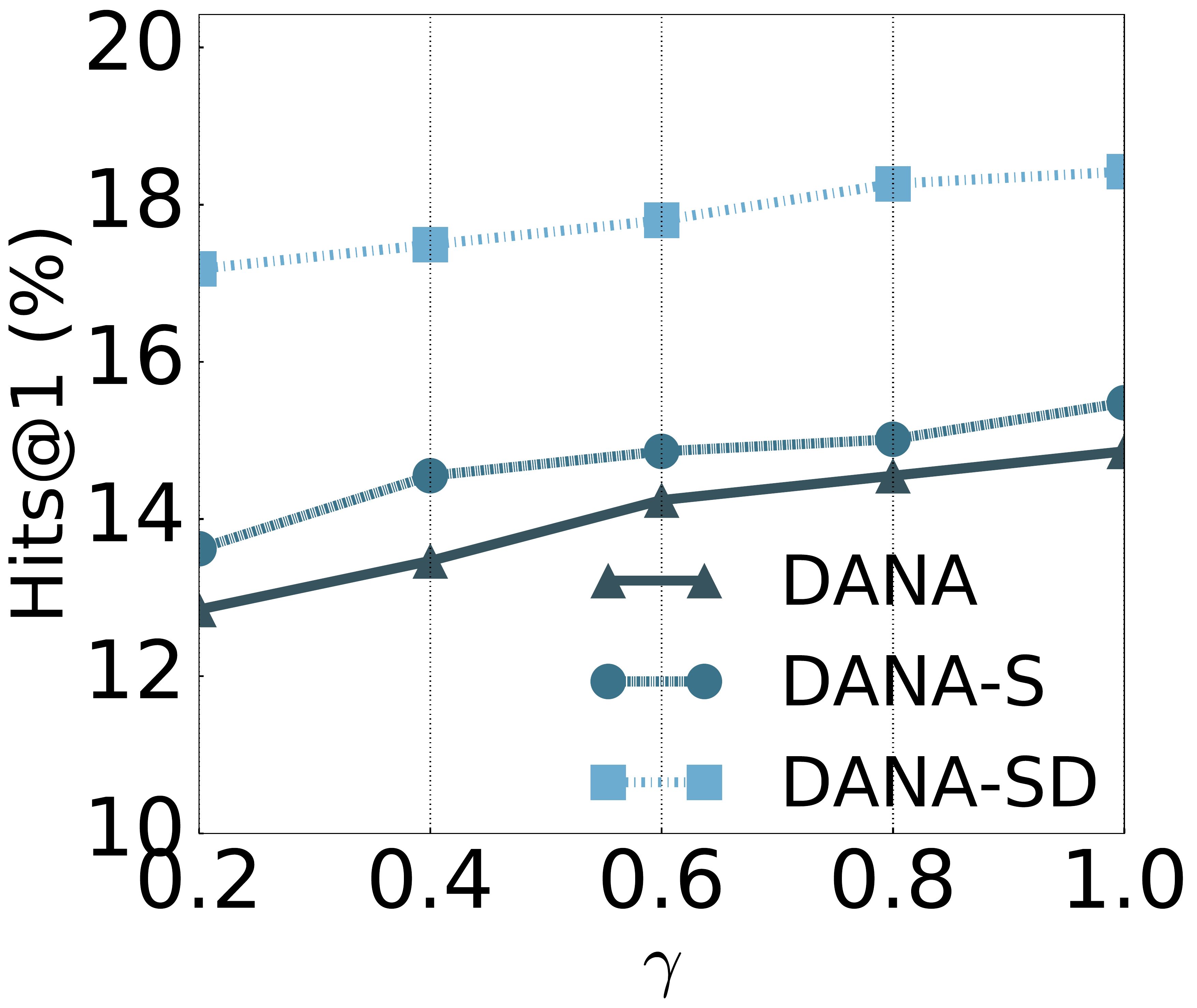}
        \label{fig:gamma_fo-tw}
    }\subfigure[Douban-Weibo]{
        \includegraphics[width=0.243\textwidth]{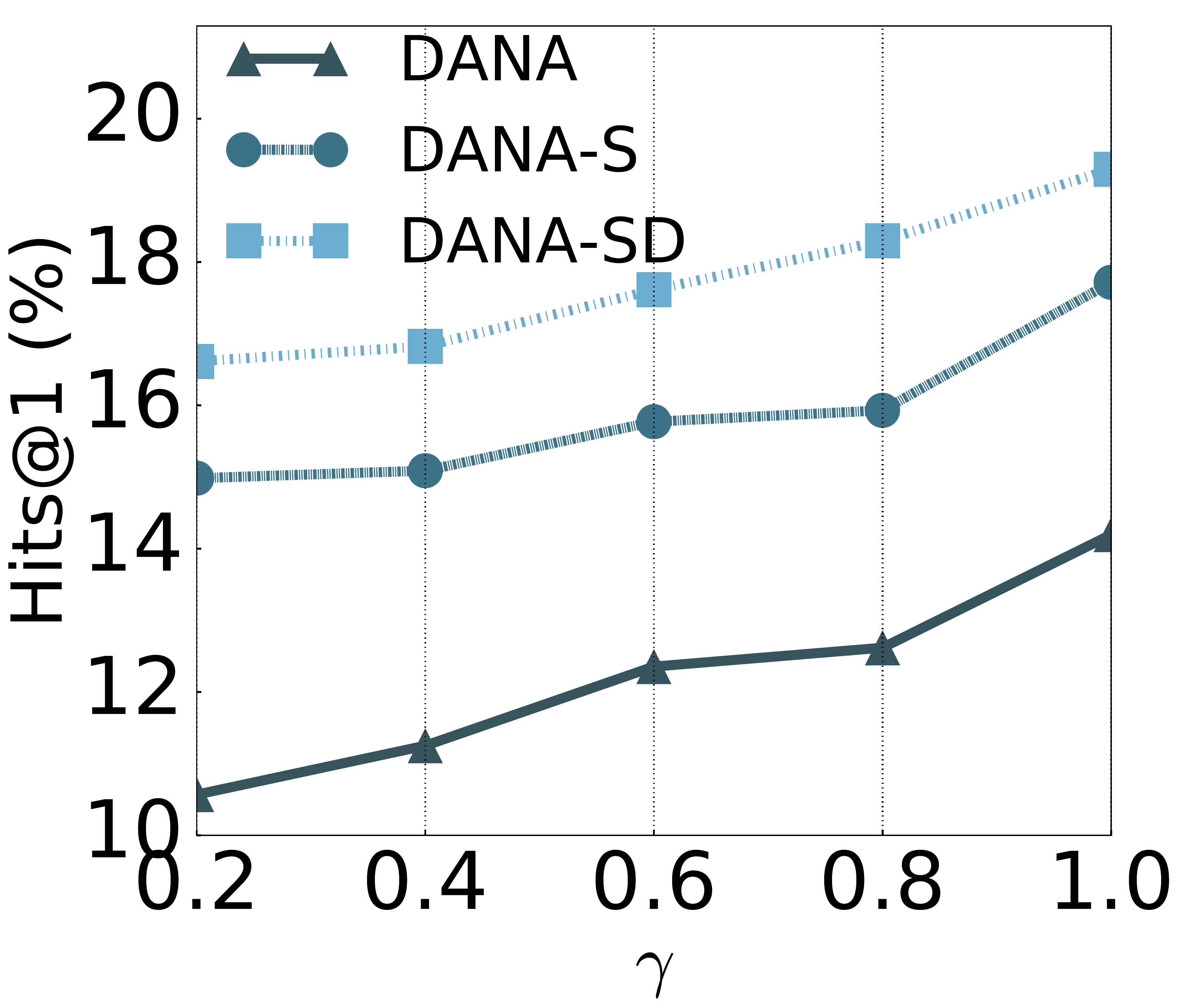}
        \label{fig:gamma_db-wb}
    }
\caption{\label{fig:gammas} Sensitivity analysis of parameter $\gamma$}
\end{figure}

Fig.\ref{fig:gammas} presents the effect of the weighting factor $\gamma$ when varying its values in $\{0.2, 0.4, 0.6, 0.8, 1.0\}$ and fixing all other parameters. The alignment performances on both Foursquare-Twitter and Douban-Weibo datasets appear an obvious increasing tend with the increase of $\gamma$, which demonstrates that the domain-adversarial learning module in DANAs plays a positive role for the alignment task.
\begin{table}[t]
    \centering
     \caption{Link Prediction Performance on Foursquare-Twitter}
    \begin{tabular}{clrrr}
    \toprule
     Dataset&	Metric(\%)	&GCN	&GCN-D	&Improve\\
    \midrule
    \multirow{4}{*}{Foursquare}&	mAP&	10.947&	12.267&	12.06\%\\
	&R@3&	8.928&	10.287&	15.22\%\\
	&R@5&	13.956&	15.862&	13.66\%\\
	&R@10&	20.367&	23.400&	14.89\%\\
	\midrule
    \multirow{4}{*}{Twitter}&	mAP&	8.651&	9.079&	4.95\%\\
	&R@3&	5.175&	5.769&	11.48\%\\
	&R@5&	8.223&	9.314&	13.27\%\\
	&R@10&	13.556&	14.979&	10.50\%\\
	\bottomrule
    \end{tabular}
    \label{tab:link_prediction_results}
\end{table}

\subsubsection{Probabilistic Design Effect}


To verify the effectiveness of our unconventional design in objective function for the alignment task, we compare MAP-based models and MSE-based models on three datasets. MAP denotes the Maximum Posterior Probability and the objective function is designed as Eq.\eqref{fun:loss_g1} in this paper. MSE denotes Minimize mean Square Error which is adopted in most of the existing distance-based approaches. In our experiments, the objective function of MSE-based alignment models is given as:
\begin{equation}
\resizebox{.91\linewidth}{!}{$
    \displaystyle
\mathcal{J}_{MSE} \!=\!\underset{(v_i^A, v_j^B)\in S}{\sum}\! \left(\Vert r_i^A - r_j^B\Vert\!-\!\frac{1}{2C}\left(\overset{C}{\underset{v_c^B}{\sum}}\Vert r_i^A - r_c^B\Vert\!+\!\overset{C}{\underset{v_c^A}{\sum}}\Vert r_c^A - r_j^B\Vert\right)\right)$}
    \label{mse-obj}
\end{equation}
where $v^A_c$ and $v^B_c$ are the negative samples. For each anchor pair, we randomly sample $C=50$ negative samples from network $A$ and network $B$ respectively. We further adapt \textit{DNA} and the distance-based model  \textit{SNNA} by replacing their objective functions with Eq.\eqref{mse-obj} and Eq.\eqref{fun:loss_g1} respectively to obtain four models for comparison, namely, (MAP-based) DNA, MSE-based DNA, (MSE-based) SNNA and MAP-based SNNA.

\begin{figure}[t]
\centering
    \subfigure[Hits@1 of DNA]{
        \includegraphics[width=0.236\textwidth]{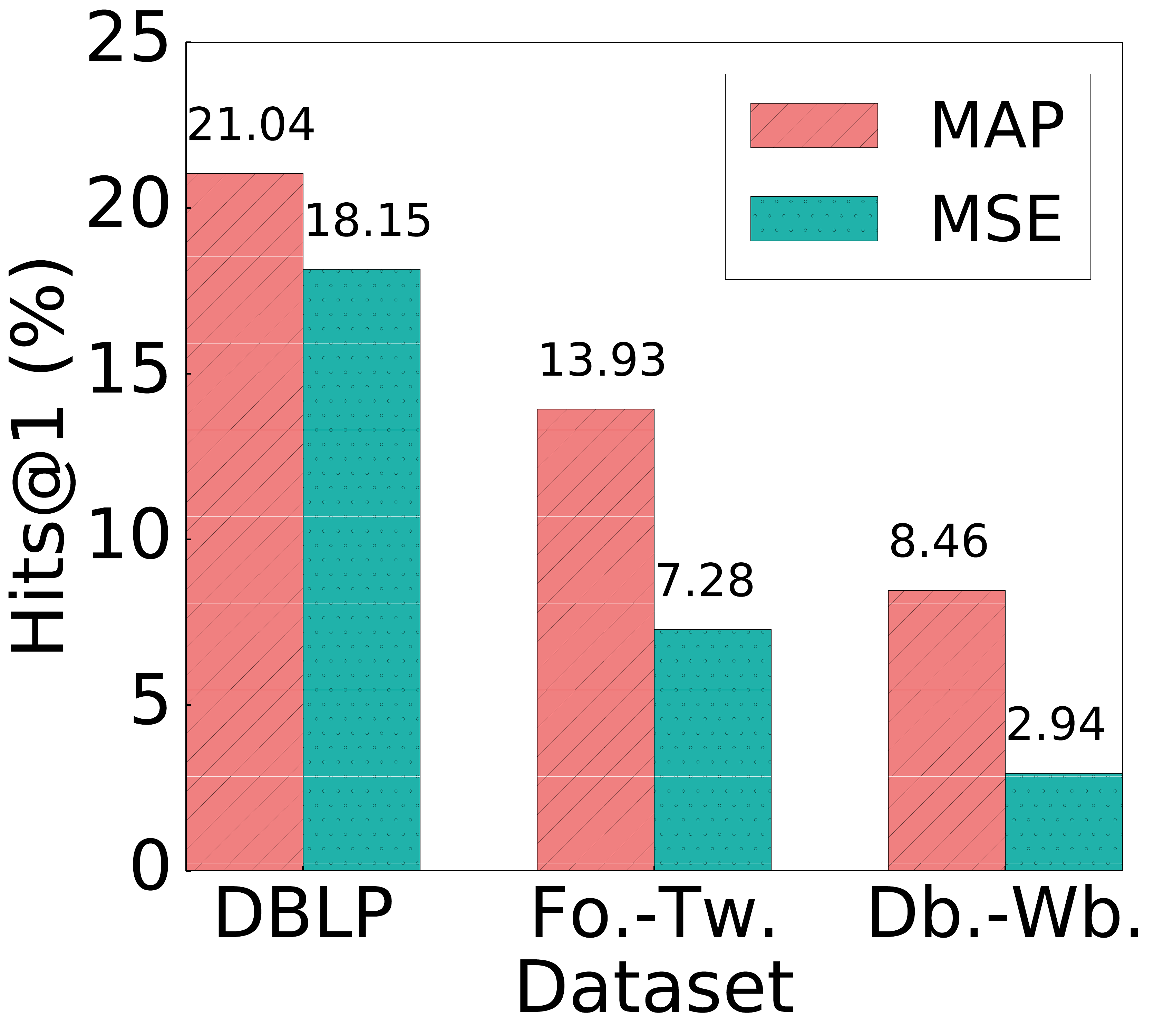}
        \label{fig:loss_DNA_hits@1}
    }\subfigure[MRR of DNA]{
        \includegraphics[width=0.236\textwidth]{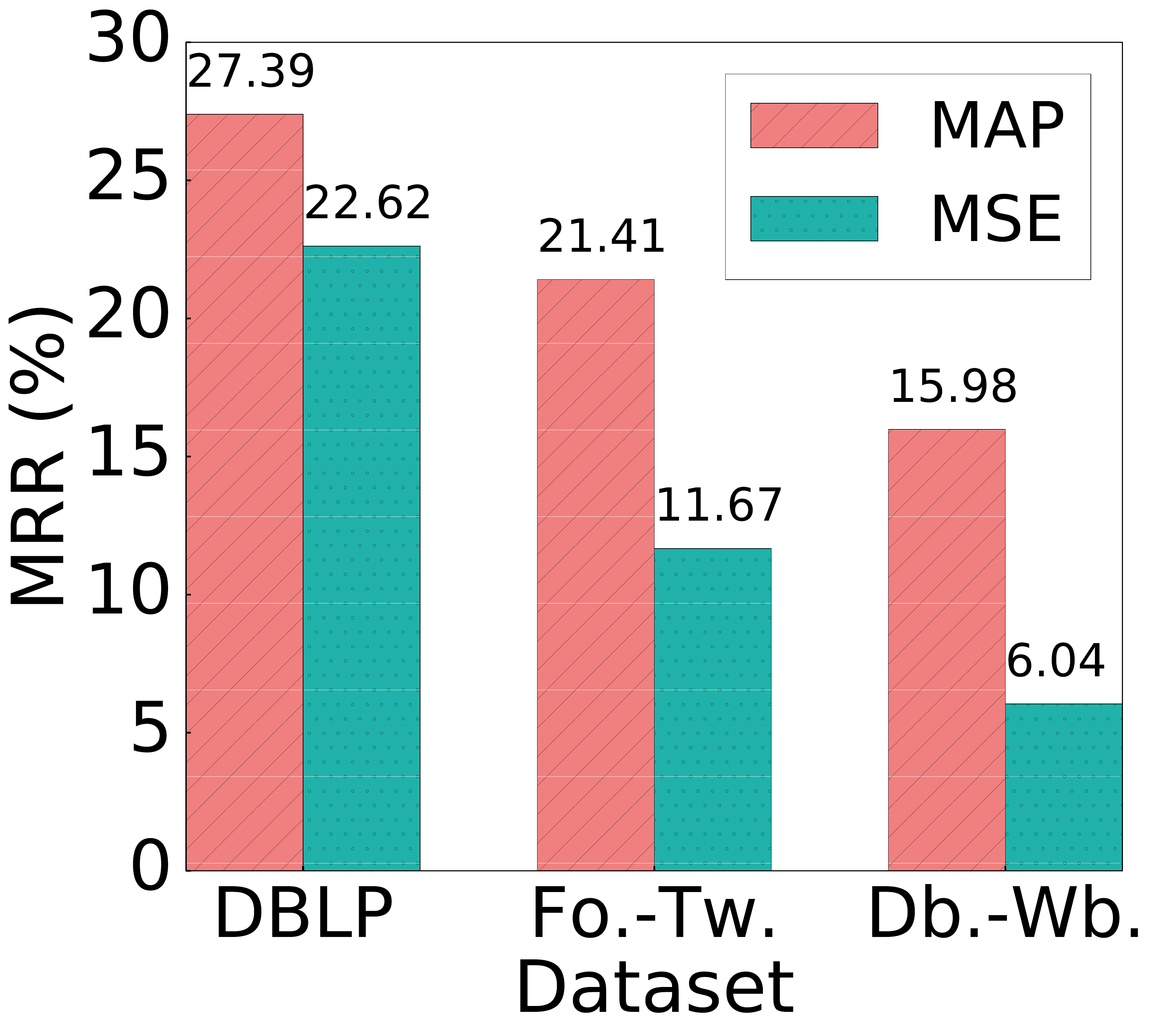}
        \label{fig:loss_DNA_mrr}
    }
    
    \subfigure[Hits@1 of SNNA]{
        \includegraphics[width=0.236\textwidth]{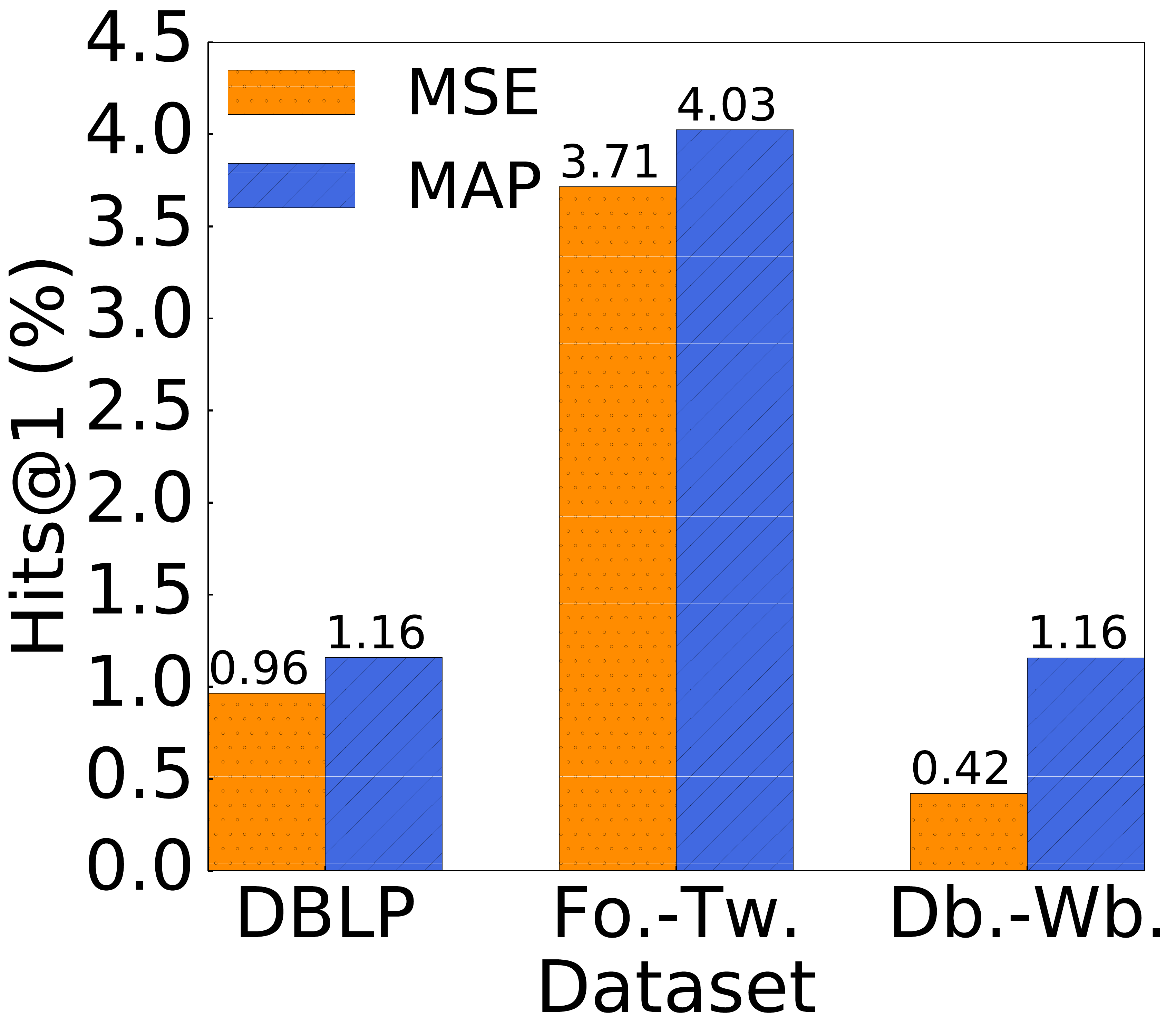}
        \label{fig:loss_SNNA_hits@1}
    }\subfigure[MRR of SNNA]{
        \includegraphics[width=0.236\textwidth]{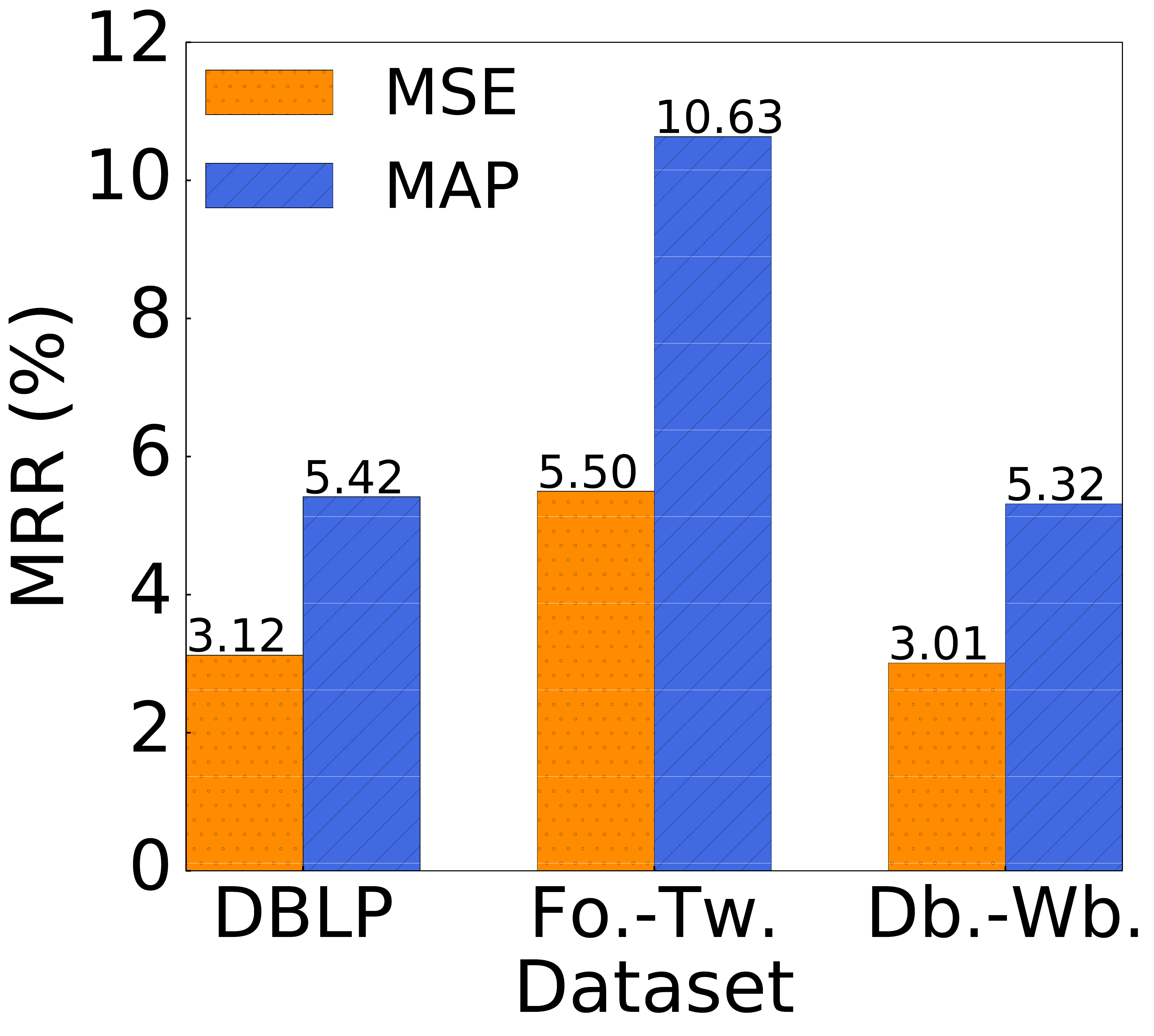}
        \label{fig:loss_SNNA_mrr}
    }
\caption{\label{fig:loss} Objective analysis of alignment task.}
\end{figure}

Fig.\ref{fig:loss_DNA_hits@1} and Fig.\ref{fig:loss_DNA_mrr} show the performance of MAP-based DNA and MSE-based DNA on three datasets. We see that \textit{DNA} lost 4.77-9.94\% MRR accuracy for the alignment when its  objective function is replaced by Eq.\eqref{mse-obj}. Fig.\ref{fig:loss_SNNA_hits@1} and Fig.\ref{fig:loss_SNNA_mrr} show the similar observation. MAP-based SNNA performs better than MSE-based SNNA on all three dataset, which illustrates the strength of our MAP-based design by viewing the alignment as a bi-directional matching problem. Note that the alignment performance of MAP-based SNNA is still much lower than that of our proposed \textit{DANAs}. One of the reasons is that the features of \textit{SNNA} learned from the network embedding may include domain-dependent signals, which cannot be eliminated in its adversarial procedure of learning the projection function between two networks. Thus, \textit{SNNA} cannot avoid domain representation bias which yields an inferior alignment performance.

\subsubsection{Directed Convolution Effect}
Recall that we propose to modify the graph convolutional network in this paper to adapt our alignment model to directed networks (See Sec.\ref{sec:directed-gcn}). To verify the effect of the directed convolution structure, we compare GCN and GCN-D ("-D" indicates an incorporation of the direction-aware convolution structure) on link prediction task within a single network, where the objective function is formulated to preserve the structural proximity \cite{tang2015line}:
\begin{equation}
    \mathcal{L} = -\underset{(i,j)\in E}{\sum}\left(\log\sigma(r_j^T \cdot r_i)+\frac{1}{C}\overset{C}{\underset{v_c\in V}{\sum}}\log\sigma(-r_c^T\cdot r_i)\right)\nonumber
\end{equation}
where $(v_i,v_c)$ denotes a negative edge randomly drawn from the noise distribution and $C$ is the number of negative edges for each observed edges $(v_i, v_j)$.

We split 90\% edges from the network for the training process. Table \ref{tab:link_prediction_results} reports the test performances of link prediction on Foursquare network and Twitter network with respect to the metrics Mean Average Precision (mAP) and Recall@k (R@k) \cite{map_and_recall}. As we expected, the performance of GCN-D all significantly improve over the conventional GCN. It implies that intentionally capturing the directions in GCNs is beneficial to the representation learning of directed networks, and in turn beneficial to the alignment of directed networks.


\subsection{Case Study: Domain-invariant Embedding}
\begin{figure}[t]
    \subfigure[DNA-S]{
        \includegraphics[width=0.357\textwidth]{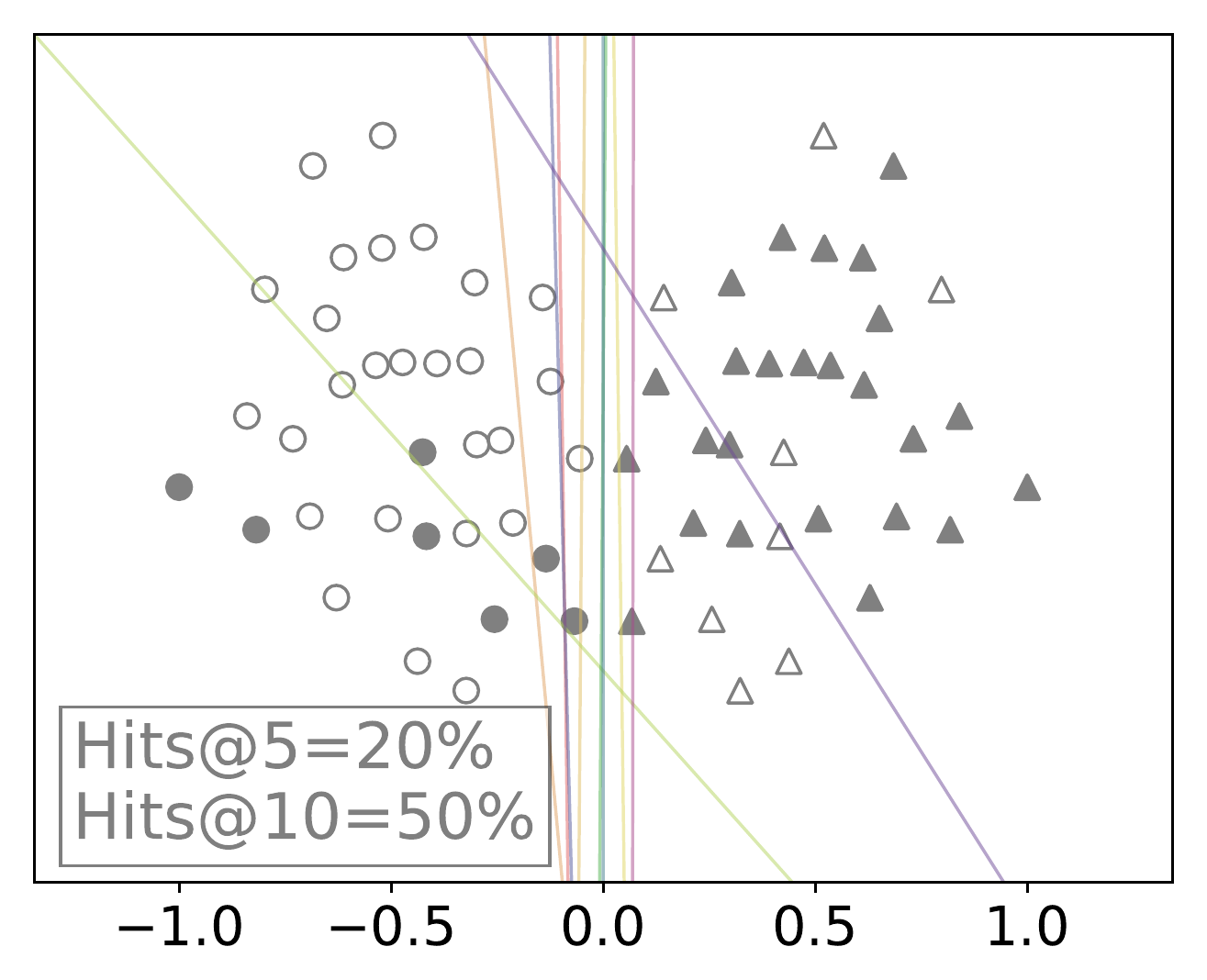}
        \label{fig:visualization_DNA}
    }
    \subfigure[DANA-S]{
        \includegraphics[width=0.357\textwidth]{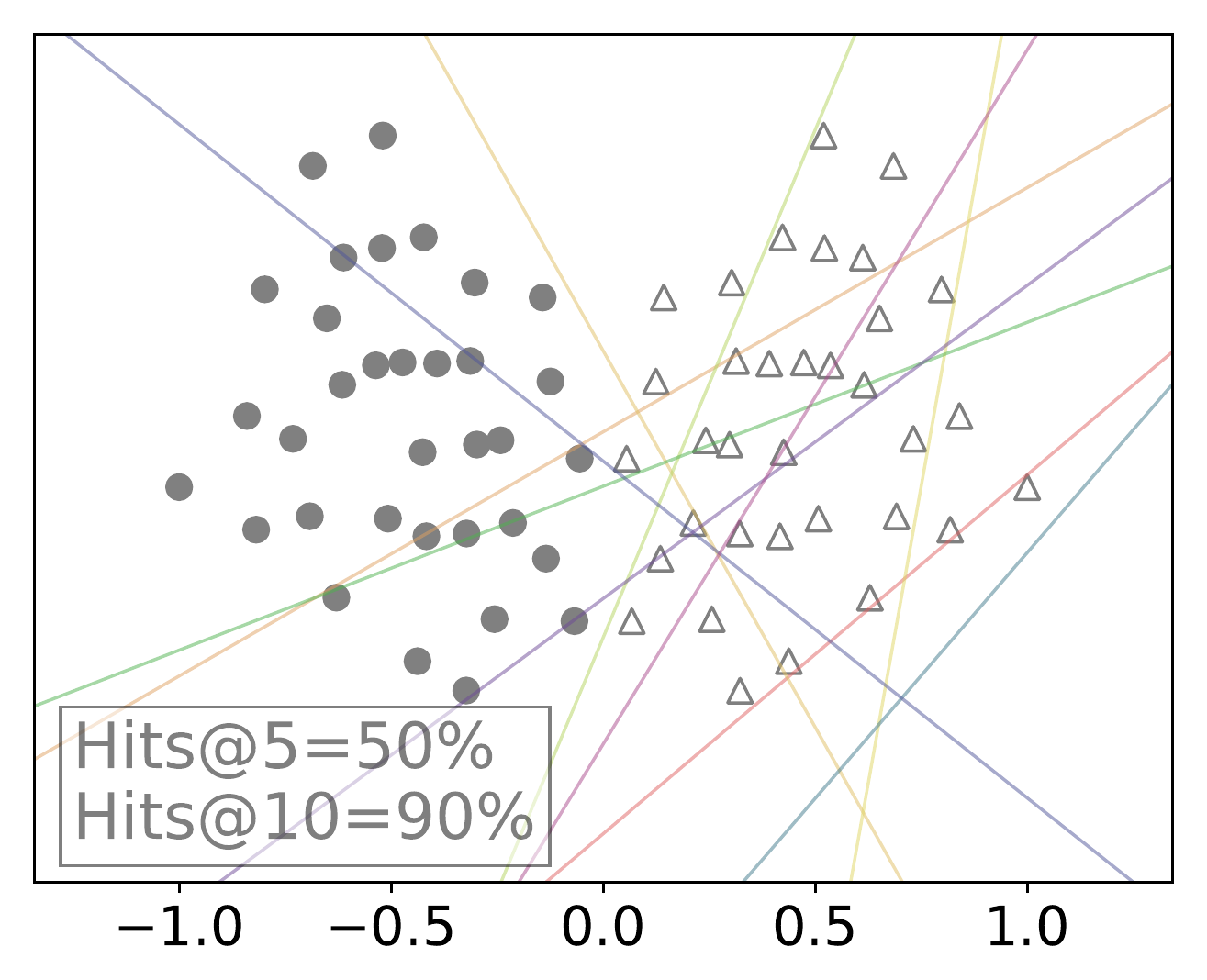}
        \label{fig:visualization_DANA}
    }
\caption{\label{fig:visualization}Hidden neuron visualization on the toy twinning-networks}
\end{figure}
To better illustrate the characteristic of our proposed model, we introduce a case study in Fig.\ref{fig:visualization} to visualize the behavior of the domain adversarial training. A twinning-networks ($N^A$ and $N^B$) is constructed as follows:
We adopt the well known Zachary's Karate network \cite{zachary1977information} as $N^A$, where the 2D embedding (coordinates) of vertices (shown as circles) are obtained via large graph layout following \cite{adai2004lgl}. (2) The nodes in $N^B$ (presented as triangles) are generated with the mirror opposite of each node in $N^A$ along the y-axis. (3) The edges of $N^B$ are generated exactly the same as that of $N^A$. (4) Each node in $N^A$ along with its corresponding node in $N^B$ are considered as an anchor in the twinning-networks.

Taking 50\% of anchors as the training set and initializing $H^A_0$ and $H^B_0$ with the coordinates,
we perform \textit{DANA-S} and \textit{DNA-S} for the alignment task with $W^A=W^B$ and $M^A=M^B$, where the network embedding module are instantiated with 1-layer GCNs. Let {\Large$\bullet$} / $\blacktriangle$ denote the points correctly classified by the domain classifier and {\Large$\circ$} / $\vartriangle$ denote the missed shot. Note that \textit{DANA-S}, integrated with domain-adversarial learning, is in pursuit of the domain invariant features, which may be not good for the domain classifier (See Fig.\ref{fig:visualization_DANA}, all nodes are classified to one domain). While the features learned with \textit{DNA-S} are domain dependent, leading to an inferior performance for the alignment task.

We visualize the weight W of the hidden neurons in the 1-layer GCNs in Fig.\ref{fig:visualization} following \cite{ganin2016domain}, where $W\in \mathbb{R}^{2\times k}$, $k=10$. Note that the neurons visualization consists of ten lines with each line corresponding to the $i$-th neuron of the hidden layer, $i=1,2,\cdots,10$. We can observe that: 
\begin{enumerate}
\item Most neurons of \textit{DNA-S} gather around and parallel to y-axis, tending to capture the discriminative feature for domain classification, since the twin-networks is y-axis symmetric.
\item \textit{DANA-S} gives a richer representation, that is, the ten lines of neurons visualization are widely dispersed.
\item The dominant pattern in the neurons visualization of \textit{DNA-S}, i.e., the lines parallel to y-axis, vanishes in that of \textit{DANA-S}, bringing a better performance for the alignment task.
\end{enumerate}

\section{Conclusion}\label{sec:conclusion}

With a conjecture that domain-dependent features hinder the network alignment performance,
we propose a representation learning-based domain-adversarial framework (\textit{DANA}) to perform network alignment, by obtaining domain-invariant representations, and develop its adaptions for specific tasks, i.e. (directed social network alignment).
Comprehensive empirical studies on three popular real-world datasets show that  \textit{DANA} can significantly improve the performance for social network alignment tasks in comparison with existing solutions.
Unlike most existing approaches which formulate the alignment task as the mapping problem between networks, Our paper triggers the discussion on the importance of feature extraction toward alignment tasks. And the proposed network alignment framework 
opens a new door to other tasks, e.g., cross-lingual knowledge graph task.


\bibliographystyle{named}
\bibliography{ijcai19-multiauthor}

\begin{thebibliography}{}

\bibitem[\protect\citeauthoryear{Adai \bgroup \em et al.\egroup
  }{2004}]{adai2004lgl}
Alex~T Adai, Shailesh~V Date, Shannon Wieland, and Edward~M Marcotte.
\newblock Lgl: creating a map of protein function with an algorithm for
  visualizing very large biological networks.
\newblock {\em Journal of molecular biology}, 340(1):179--190, 2004.

\bibitem[\protect\citeauthoryear{Cao and Yu}{2016}]{cao2016asnets}
Xuezhi Cao and Yong Yu.
\newblock Asnets: A benchmark dataset of aligned social networks for
  cross-platform user modeling.
\newblock In {\em Proceedings of the 25th ACM International on Conference on
  Information and Knowledge Management}, pages 1881--1884. ACM, 2016.

\bibitem[\protect\citeauthoryear{Dai \bgroup \em et al.\egroup
  }{2018}]{dai2018adversarial}
Quanyu Dai, Qiang Li, Jian Tang, and Dan Wang.
\newblock Adversarial network embedding.
\newblock In {\em Thirty-Second AAAI Conference on Artificial Intelligence},
  2018.

\bibitem[\protect\citeauthoryear{Defferrard \bgroup \em et al.\egroup
  }{2016}]{defferrard2016convolutional}
Micha{\"e}l Defferrard, Xavier Bresson, and Pierre Vandergheynst.
\newblock Convolutional neural networks on graphs with fast localized spectral
  filtering.
\newblock In {\em Advances in neural information processing systems}, pages
  3844--3852, 2016.

\bibitem[\protect\citeauthoryear{Ganin \bgroup \em et al.\egroup
  }{2016}]{ganin2016domain}
Yaroslav Ganin, Evgeniya Ustinova, Hana Ajakan, Pascal Germain, Hugo
  Larochelle, Fran{\c{c}}ois Laviolette, Mario Marchand, and Victor Lempitsky.
\newblock Domain-adversarial training of neural networks.
\newblock {\em The Journal of Machine Learning Research}, 17(1):2096--2030,
  2016.

\bibitem[\protect\citeauthoryear{Goodfellow \bgroup \em et al.\egroup
  }{2014}]{goodfellow2014generative}
Ian Goodfellow, Jean Pouget-Abadie, Mehdi Mirza, Bing Xu, David Warde-Farley,
  Sherjil Ozair, Aaron Courville, and Yoshua Bengio.
\newblock Generative adversarial nets.
\newblock In {\em Advances in neural information processing systems}, pages
  2672--2680, 2014.

\bibitem[\protect\citeauthoryear{Jean \bgroup \em et al.\egroup
  }{2014}]{jean2014using}
S{\'e}bastien Jean, Kyunghyun Cho, Roland Memisevic, and Yoshua Bengio.
\newblock On using very large target vocabulary for neural machine translation.
\newblock {\em arXiv preprint arXiv:1412.2007}, 2014.

\bibitem[\protect\citeauthoryear{Kipf and Welling}{2016}]{kipf2016semi}
Thomas~N Kipf and Max Welling.
\newblock Semi-supervised classification with graph convolutional networks.
\newblock {\em arXiv preprint arXiv:1609.02907}, 2016.

\bibitem[\protect\citeauthoryear{Li \bgroup \em et al.\egroup
  }{2018}]{DBLP:conf/aaai/LiHW18}
Qimai Li, Zhichao Han, and Xiao{-}Ming Wu.
\newblock Deeper insights into graph convolutional networks for semi-supervised
  learning.
\newblock In {\em Proceedings of the Thirty-Second {AAAI} Conference on
  Artificial Intelligence, (AAAI-18), the 30th innovative Applications of
  Artificial Intelligence (IAAI-18), and the 8th {AAAI} Symposium on
  Educational Advances in Artificial Intelligence (EAAI-18), New Orleans,
  Louisiana, USA, February 2-7, 2018}, pages 3538--3545, 2018.

\bibitem[\protect\citeauthoryear{Li \bgroup \em et al.\egroup }{2019}]{SNNA}
Chaozhuo Li, Yukun Wang, Senzhang Wang, Yun Liu, Philip Yu, Zhoujun Li, and
  Yanbo Liang.
\newblock Adversarial learning for weakly-supervised social network alignment.
\newblock In {\em Thirty-Third AAAI Conference on Artificial Intelligence},
  2019.

\bibitem[\protect\citeauthoryear{Liu \bgroup \em et al.\egroup
  }{2016}]{liu2016aligning}
Li~Liu, William~K Cheung, Xin Li, and Lejian Liao.
\newblock Aligning users across social networks using network embedding.
\newblock In {\em International Joint Conference on Artificial Intelligence},
  pages 1774--1780, 2016.

\bibitem[\protect\citeauthoryear{Man \bgroup \em et al.\egroup
  }{2016}]{man2016predict}
Tong Man, Huawei Shen, Shenghua Liu, Xiaolong Jin, and Xueqi Cheng.
\newblock Predict anchor links across social networks via an embedding
  approach.
\newblock In {\em International Joint Conference on Artificial Intelligence},
  volume~16, pages 1823--1829, 2016.

\bibitem[\protect\citeauthoryear{Mu \bgroup \em et al.\egroup
  }{2016}]{Mu:2016:UIL:2939672.2939849}
Xin Mu, Feida Zhu, Ee-Peng Lim, Jing Xiao, Jianzong Wang, and Zhi-Hua Zhou.
\newblock User identity linkage by latent user space modelling.
\newblock In {\em Proceedings of the 22nd ACM SIGKDD International Conference
  on Knowledge Discovery and Data Mining}, pages 1775--1784. ACM, 2016.

\bibitem[\protect\citeauthoryear{Pan \bgroup \em et al.\egroup
  }{2010}]{panjialinWWW10}
Sinno~Jialin Pan, Xiaochuan Ni, Jian-Tao Sun, Qiang Yang, and Zheng Chen.
\newblock Cross-domain sentiment classification via spectral feature alignment.
\newblock In {\em The 19th International World Wide Web Conference}, pages
  751--760. ACM, 2010.

\bibitem[\protect\citeauthoryear{Perozzi \bgroup \em et al.\egroup
  }{2014}]{Perozzi2014DeepWalk}
Bryan Perozzi, Rami Al-Rfou, and Steven Skiena.
\newblock Deepwalk: Online learning of social representations.
\newblock In {\em Acm Sigkdd International Conference on Knowledge Discovery \&
  Data Mining}, 2014.

\bibitem[\protect\citeauthoryear{Radev \bgroup \em et al.\egroup
  }{2002}]{radev2002evaluating}
Dragomir~R Radev, Hong Qi, Harris Wu, and Weiguo Fan.
\newblock Evaluating web-based question answering systems.
\newblock In {\em Proceedings of the Third International Conference on Language
  Resources and Evaluation, {LREC} 2002, May 29-31, 2002, Las Palmas, Canary
  Islands, Spain}, 2002.

\bibitem[\protect\citeauthoryear{Schlichtkrull \bgroup \em et al.\egroup
  }{2018}]{schlichtkrull2018modeling}
Michael Schlichtkrull, Thomas~N Kipf, Peter Bloem, Rianne Van Den~Berg, Ivan
  Titov, and Max Welling.
\newblock Modeling relational data with graph convolutional networks.
\newblock In {\em European Semantic Web Conference}, pages 593--607. Springer,
  2018.

\bibitem[\protect\citeauthoryear{Tan \bgroup \em et al.\egroup
  }{2014}]{tan2014mapping}
Shulong Tan, Ziyu Guan, Deng Cai, Xuzhen Qin, Jiajun Bu, and Chun Chen.
\newblock Mapping users across networks by manifold alignment on hypergraph.
\newblock In {\em Twenty-Eighth AAAI Conference on Artificial Intelligence},
  volume~14, pages 159--165, 2014.

\bibitem[\protect\citeauthoryear{Tang \bgroup \em et al.\egroup
  }{2008}]{tang2008arnetminer}
Jie Tang, Jing Zhang, Limin Yao, Juanzi Li, Li~Zhang, and Zhong Su.
\newblock Arnetminer: extraction and mining of academic social networks.
\newblock In {\em Proceedings of the 14th ACM SIGKDD international conference
  on Knowledge discovery and data mining}, pages 990--998. ACM, 2008.

\bibitem[\protect\citeauthoryear{Tang \bgroup \em et al.\egroup
  }{2015}]{tang2015line}
Jian Tang, Meng Qu, Mingzhe Wang, Ming Zhang, Jun Yan, and Qiaozhu Mei.
\newblock Line: Large-scale information network embedding.
\newblock In {\em Proceedings of the 24th international conference on world
  wide web}, pages 1067--1077, 2015.

\bibitem[\protect\citeauthoryear{Wang \bgroup \em et al.\egroup
  }{2017}]{wang2017irgan}
Jun Wang, Lantao Yu, Weinan Zhang, Yu~Gong, Yinghui Xu, Benyou Wang, Peng
  Zhang, and Dell Zhang.
\newblock Irgan: A minimax game for unifying generative and discriminative
  information retrieval models.
\newblock In {\em Proceedings of the 40th International ACM SIGIR conference on
  Research and Development in Information Retrieval}, pages 515--524. ACM,
  2017.

\bibitem[\protect\citeauthoryear{Wang \bgroup \em et al.\egroup
  }{2018}]{wang2018graphgan}
Hongwei Wang, Jia Wang, Jialin Wang, Miao Zhao, Weinan Zhang, Fuzheng Zhang,
  Xing Xie, and Minyi Guo.
\newblock Graphgan: Graph representation learning with generative adversarial
  nets.
\newblock In {\em Thirty-Second AAAI Conference on Artificial Intelligence},
  2018.

\bibitem[\protect\citeauthoryear{Weiss \bgroup \em et al.\egroup
  }{2016}]{weiss2016survey}
Karl Weiss, Taghi~M Khoshgoftaar, and DingDing Wang.
\newblock A survey of transfer learning.
\newblock {\em Journal of Big Data}, 3(1):9, 2016.

\bibitem[\protect\citeauthoryear{{Wikipedia
  contributors}}{2019}]{map_and_recall}
{Wikipedia contributors}.
\newblock Evaluation measures (information retrieval) --- {Wikipedia}{,} the
  free encyclopedia.
\newblock
  \url{https://en.wikipedia.org/w/index.php?title=Evaluation_measures_(information_retrieval)&oldid=889157178},
  2019.
\newblock [Online; accessed 21-May-2019].

\bibitem[\protect\citeauthoryear{Xie \bgroup \em et al.\egroup
  }{2017}]{xie2017controllable}
Qizhe Xie, Zihang Dai, Yulun Du, Eduard Hovy, and Graham Neubig.
\newblock Controllable invariance through adversarial feature learning.
\newblock In {\em Advances in Neural Information Processing Systems}, pages
  585--596, 2017.

\bibitem[\protect\citeauthoryear{Zachary}{1977}]{zachary1977information}
Wayne~W Zachary.
\newblock An information flow model for conflict and fission in small groups.
\newblock {\em Journal of anthropological research}, 33(4):452--473, 1977.

\bibitem[\protect\citeauthoryear{Zhang and Philip}{2015}]{zhang2015integrated}
Jiawei Zhang and S~Yu Philip.
\newblock Integrated anchor and social link predictions across social networks.
\newblock In {\em International Joint Conference on Artificial Intelligence},
  pages 2125--2132, 2015.

\bibitem[\protect\citeauthoryear{Zhang \bgroup \em et al.\egroup
  }{2017}]{zhang2017adversarial}
Yizhe Zhang, Zhe Gan, Kai Fan, Zhi Chen, Ricardo Henao, Dinghan Shen, and
  Lawrence Carin.
\newblock Adversarial feature matching for text generation.
\newblock {\em arXiv preprint arXiv:1706.03850}, 2017.

\end{thebibliography}

\end{document}